\documentclass[11pt]{article} 

\usepackage[numbers]{natbib}
\usepackage{bm}
\usepackage{amsmath}
\usepackage{amssymb}
\usepackage{graphicx} 
\usepackage{subfig}

\DeclareMathOperator{\oec}{\mathbf{o\mkern-3mu e}}

\DeclareMathOperator{\doe}{\delta\mkern-1mu\mathbf{o\mkern-3mu e}}

\begin{document}

\title{Geometric Perspectives on Fundamental Solutions in the Linearized Satellite Relative Motion Problem}                      
%
%

\author{Ethan Burnett\thanks{PhD Candidate, Aerospace Engineering Sciences, University of Colorado Boulder, \texttt{ethan.burnett@colorado.edu}} \ and Hanspeter Schaub\thanks{Glenn L. Murphy Chair of Engineering, Smead Department of Aerospace Engineering Sciences, University of Colorado Boulder}%
}


\date{}
\maketitle

\begin{abstract}
Understanding natural relative motion trajectories is critical to enable fuel-efficient multi-satellite missions operating in complex environments. This paper studies the problem of computing and efficiently parameterizing satellite relative motion solutions for linearization about a closed chief orbit. By identifying the analytic relationship between Lyapunov-Floquet transformations of the relative motion dynamics in different coordinate systems, new means are provided for rapid computation and exploration of the types of close-proximity natural relative motion available in various applications. The approach is demonstrated for the Keplerian relative motion problem with general eccentricities in multiple coordinate representations. The Keplerian assumption enables an analytic approach, leads to new geometric insights, and allows for comparison to prior linearized relative motion solutions.
\end{abstract}

\section{Introduction}
A large body of work has studied the nature of satellite relative motion in Keplerian and perturbed orbits. Generally, linearized or nonlinear dynamic models tailored to a particular application are derived and solved to yield linear or nonlinear solutions. The problem of close-proximity relative motion in the vicinity of satellites in circular orbits was solved by Clohessy and Wiltshire in 1960 \cite{clohessy1},  and the solution for general elliptical orbits was derived by Tschauner and Hempel in 1965 \cite{tschauner1}. Both of these solutions provide analytic and reasonably concise expressions for solving the close-proximity relative motion problem. In the years since, there has been a great deal of continued analysis of the Keplerian relative motion problem, focusing on relative motion design using the natural solutions and impulsive maneuvers, or continuous feedback control on the relative state. Additionally, many researchers have explored the influence of orbital perturbations on the relative motion dynamics and their solutions \cite{KechAngVel1998, MeanElem, Casotto}. In the course of this research, many different coordinate representations of the relative motion problem have been developed for various divergent applications. The most common types are local coordinate representations, in which the relative state is resolved in terms of local relative position and velocity, orbit element differences \cite{MeanElemImpulsive}, and relative orbit elements (ROEs) \cite{schaub, DAmico_EccInc}. ROEs in particular are of note because they describe relative motion similarly to orbit elements, but with more of a local geometric interpretation. While many works pursue developments in only a singe coordinate representation, there have been some successful works that exploit the relationship between coordinate representations for achieving control and modeling goals \cite{LocalCart}.

One interesting line of work is in determining the simplest and most convenient parameterizations of natural relative motion. Largely a question of the choice of coordinates, this also involves factoring the resultant solution in a given set of coordinates in a manner convenient or illuminating for the astrodynamicist. The state transition matrix is an unwieldy means by which to explore relative motion. Instead, other sets of fundamental linear solutions can be chosen to serve as a functional basis. Custom geometric interpretations of the solutions might also be possible, in which the relative motion solution is factored into a more concise or workable form. One example of this is the nonsingular relative orbit element set for the Clohessy-Wiltshire solution \cite{schaub, Bennett2016c}. The pursuit of desirable parameterizations of relative motion is a solved problem for the Clohessy-Wiltshire case, a manageable problem for more general elliptic orbits, but is largely unexplored for the wide variety of periodic and almost-periodic orbits in multi-body and small-body applications. In particular, to maximize geometric insight of natural relative motion in these more complex settings, various types of relative state coordinates must be explored. Some of these might be altogether different from the more familiar types used in Keplerian or weakly perturbed orbits. 

\begin{figure}[h!]
\centering
\includegraphics[]{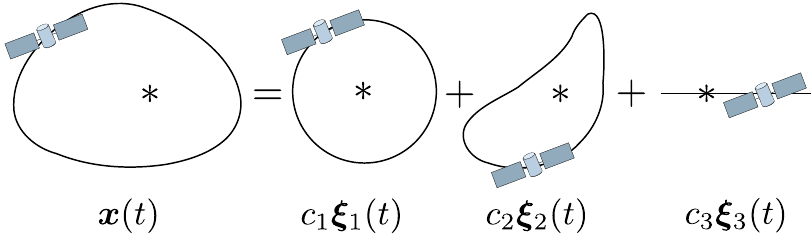}
\caption{Example Satellite Relative Motion as a Sum of Individual Modes}
\label{fig:ModalConc}
\end{figure}

To address the problem of efficiently studying the nature of relative motion in the vicinity of general closed orbits, this paper leverages the classical idea of the modal decomposition, used extensively in the theory of vibrations \cite{meirovitch2001fundamentals}. All small deflections of a continuous and homogeneous body can be expressed as a linear weighted sum of independent mode shapes, which each have their own associated frequency. In the same manner for the satellite relative motion problem in the vicinity of a closed orbit, all possible motions are the sum of 6 independent fundamental motions with their own shapes and associated frequencies. Figure \ref{fig:ModalConc} illustrates this conceptually with a depiction of relative motion decomposed into three simpler constituent modal motions. One benefit of the modal decomposition approach is that oscillatory, unstable, and drift motions are naturally isolated from one another. There are also many other benefits which will be discussed.

Several classical concepts and past works are highly relevant to this paper. First, this paper employs Lyapunov-Floquet theory \cite{Nayfeh:1979uq}, whose application to the relative motion problem has seen limited study thus far. In Reference~\citenum{Sherrill2015LyapunovFloquet}, a Lyapunov-Floquet (LF) transformation relating the Clohessy-Wiltshire and Tschauner Hempel dynamics is exploited for relative motion control design. In Reference~\citenum{OgundeleLFConference}, the authors apply the LF transformation to the Tschauner-Hempel problem with cubic nonlinearities and examine the effect of the nonlinearities on the dynamical variables via averaging theory. Additionally, a simple LF transformation is used in Reference~\citenum{KoenigJ2} when incorporating the secular effect of the $J_{2}$ perturbation on orbit element differences. The LF transformation is applied to control design elsewhere in literature. Reference~\citenum{Dshmukh_Sinha_LFControl} studies control design for dynamical systems with time-periodic coefficients using the LF transformation and the backstepping technique. This is applied to control of a system with two statically coupled pendula subject to periodic forcing. In Reference~\citenum{Montagnier_LFControl}, control of systems with periodic coefficients is discussed and LF theory is applied to control of an industrial mechanism. In addition to past work applying LF theory, there has been some work on geometric analysis of relative motion solutions for eccentric orbits. Reference \citenum{SinclairGeometricTH} explores the geometric interpretation of the Tschauner-Hempel \cite{tschauner1} and Yamanaka-Ankerson \cite{yamanaka1} solutions, as well as the novel geometric parameterizations of those solutions. Some of the fundamental solutions derived in that paper are analogous to the ones obtained in this paper in the analytic application to Keplerian relative motion dynamics. Finally, this paper also makes use of the geometric method relating local curvilinear or Cartesian coordinate representations to orbit element differences \cite{LocalCart, gim1}. 

Most past applications of the LF transformation to satellite relative motion and other systems have been explicitly for control design and not for relative motion planning, and past developments of fundamental linear relative motion solutions and ROEs have not made extensive use of Lyapunov-Floquet theory. Furthermore, the relationship between LF transformations in various coordinates has not been appreciated in literature studying the Keplerian relative motion problem. This paper fills that gap in the literature, encouraging works using multiple sets of coordinates to take greater advantage of LF theory. It also highlights the benefits of examining the relative motion problem from the perspective of fundamental relative motion modes.

This paper stems from a larger research effort to describe spacecraft relative motion in the vicinity of almost-periodic orbits through a highly generalized application of the modal decomposition method \cite{Burnett2020_Ryugu}. The authors are exploring the numerical application of the LF transformation to study relative motion in the vicinity of desirable long-term stable orbits in highly perturbed environments, such as around asteroids and in multi-body systems. The nature of relative motion in such orbits is examined through a perturbative application of Lyapunov-Floquet theory to the linearized relative motion problem. 
In essence, the relative state is approximately described for some timespan by a modal decomposition, by discarding the small non-periodic part of the system plant matrix.
Formation and rendezvous design is highly simplified in such settings if the underlying approximate relative motion modes can be identified. They serve as a simple basis for admissible natural motions, which can be numerically re-computed as needed to achieve the necessary accuracy. In the application of the necessary numerical methods, it has been numerically more convenient to work in orbit element differences than local Cartesian or curvilinear coordinates, even though the latter coordinates are much more directly useful for analysis, relative motion design, and control. As a result, the relationship of the LF transformations obtained in different relative motion coordinate sets became a topic of interest. This work takes several steps back from numerical efforts with perturbed dynamics and examines the application of the modal decomposition to the Keplerian relative motion problem to seek analytical insight into these relative motion modes.

This work is organized as follows. First, in Section 2, the Lyapunov-Floquet (LF) transformation of the relative motion problem is reviewed, for which the relative state is transformed by an orbit-periodic linear transformation to a coordinate set with linear time-invariant (LTI) dynamics. The idea of the modal decomposition is also introduced. Afterwards, the relationship between LF transformations and LTI forms for any two sets of relative motion coordinates is derived. The general dynamics of relative motion in local coordinates and quasi-nonsingular (QNS) orbit element differences are discussed, and the LF transformation and LTI form for the Keplerian case is then derived in QNS element differences in Section 3. This facilitates the derivation of the LF transformations and LTI forms for local Cartesian and spherical coordinate representations. This exercise provides some new insights, connections to past works, and solutions with useful properties. Section 4 briefly discusses the application of the modal decomposition method to perturbed dynamics. Overall, the paper provides an illuminating and unifying approach to parameterizing and exploring relative motion in the vicinity of closed orbits. 

\section{Fundamentals and Theory}
\subsection{Motivating Example}
To motivate the arguments in this paper, consider an introductory exercise using the simple well-known relative motion problem defined by Clohessy and Wiltshire \cite{clohessy1}. This problem studies the dynamics of the relative state of a \textit{deputy} spacecraft with respect to another spacecraft called the \textit{chief}, which is in a circular orbit. The relative state is augmented relative position and velocity $\bm{x} = \left(\bm{\rho}^{\top}, \ \bm{\rho}'^{\top}\right)^{\top}$  resolved in a chief-centered rotating coordinate frame called the Hill or local vertical-local horizontal (LVLH) frame as below:
\begin{subequations}
\label{relpos1}
\begin{align}
\bm{\rho} = & \  x\hat{\bm{e}}_{r} + y\hat{\bm{e}}_{t} + z\hat{\bm{e}}_{n} \\
\bm{\rho}' = & \ \frac{^{H}\text{d}}{\text{d}t}\left(\bm{\rho}\right) = \dot{x}\hat{\bm{e}}_{r} + \dot{y}\hat{\bm{e}}_{t} + \dot{z}\hat{\bm{e}}_{n}
\end{align}
\end{subequations}
The vector triad $\left\{ \hat{\bm{e}}_{r}, \ \hat{\bm{e}}_{t}, \ \hat{\bm{e}}_{n}\right\}$ forming the LVLH frame is defined below in terms of the chief inertial position, velocity, and orbit angular momentum vectors $\bm{r}_{c}$, $\bm{v}_{c}$, and $\bm{h}_{c}$, and $(\ )'$ denotes the time derivative of quantities as seen in this frame.
\begin{subequations}
\label{relpos1}
\begin{align}
\hat{\bm{e}}_{r} = & \ \bm{r}_{c}/r_{c} \\
\hat{\bm{e}}_{n} = & \ \bm{h}_{c}/h_{c} \\
\hat{\bm{e}}_{t} = & \ \hat{\bm{e}}_{n} \times\hat{\bm{e}}_{r} 
\end{align}
\end{subequations}
The linearized unforced relative motion dynamics for the Clohessy Wiltshire (CW) problem are given below:
\begin{subequations}
\begin{align}
\ddot{x} - 2n\dot{y} - 3n^{2}x = 0 \\
\ddot{y} + 2n\dot{x} = 0 \\
\ddot{z} + n^{2}z = 0
\end{align}
\end{subequations}
where $n = \sqrt{\mu/a^{3}}$ is the mean motion. Note that the out-of-plane $z$ motion is a simple harmonic oscillator. The solutions to any linearized relative motion equations can generally given in an STM format as $\bm{x}(t) = [\Phi(t,t_{0})]\bm{x}(t_{0})$. Ignoring the simple and decoupled $z$ component of the solution, the planar part of the STM is given below with epoch time $t_{0} = 0$:
\begin{equation}
\label{CWstm2D}
[\Phi(t)] = \begin{bmatrix} (4 - 3\cos{nt}) & 0 & \frac{\sin{nt}}{n}  & \frac{2}{n}(1 - \cos{nt}) \\
6(\sin{nt} - nt) & 1 & -\frac{2}{n}(1-\cos{nt}) & \frac{4}{n}\sin{nt} - 3t \\
3n\sin{nt} & 0 & \cos{nt} & 2\sin{nt} \\
-6n(1-\cos{nt}) & 0 & -2\sin{nt} & 4\cos{nt} - 3
\end{bmatrix}
\end{equation}

To illuminate the nature and types of planar relative motion permitted by Eq. \eqref{CWstm2D}, there are a few options. First, in the case of CW dynamics, the first two rows of the 4-state STM in Eq. \eqref{CWstm2D} can be factored into a simple and geometrically insightful pair of expressions:
\begin{subequations}
\label{XandY_CW}
\begin{align}
x(t) = & \ A_{0}\cos{(nt + \alpha)} + x_{\text{off}} \\ 
y(t) = & \ -2A_{0}\sin{(nt + \alpha)} -\frac{3}{2}nt x_{\text{off}} + y_{\text{off}}
\end{align}
\end{subequations}
where $A_{0}$, $\alpha$, $x_{\text{off}}$, and $y_{\text{off}}$ are ROEs that are functions of the initial relative state conditions. These are defined in Reference \citenum{schaub}. Eq. \eqref{XandY_CW} shows that the planar relative motion is in a 2:1 ellipse when $x_{\text{off}} = 0$, and otherwise drifts in the along-track direction. This concise and highly specialized expression stems from the simplicity of the CW dynamics. In relative motion cases where the STM is more complicated than the form given in Eq. \eqref{CWstm2D}, an alternate and more general approach for understanding the relative motion is needed.

One alternate approach is to consider an expression of the relative motion in terms of fundamental solutions~$\bm{\xi}_{i}$: 
\begin{equation}
\label{FS_rm1}
\bm{x}(t) = \sum_{i = 1}^{6}c_{i}\bm{\xi}_{i}(t)
\end{equation}
The $c_{i}$ constants are functions of the initial conditions. A prudent choice of fundamental solutions enables the relative motion to be investigated and designed by simply varying the weighing constants, with the fundamental solutions designed such that their geometry is as simple as possible. In this manner, the constants perform a similar function to ROEs, by directly providing geometric insight.

The most obvious fundamental solutions are the columns of the STM, for which $\bm{c} = \bm{x}_{0}$ in Eq. \eqref{FS_rm1}. These are typically inconvenient for geometric interpretation. For the case of the CW problem, two of the four columns of the planar STM given by Eq. \eqref{CWstm2D} have drifting components, whereas the drifting part of the solution is one-dimensional. A superior parameterization would thus isolate the drifting motion to only one fundamental solution, with the associated constant $c_{i}$ providing a no-drift constraint $c_{i} = 0$. Such a set of solutions is offered by the eigenvalue decomposition of the planar CW problem into independent modes. More generally, the modal decomposition serves as an attractive parameterization of the relative motion solution regardless of the dynamics and orbit geometry, for any periodic orbit. This will be discussed later.

\begin{figure*}[!b]
\centering
\includegraphics[]{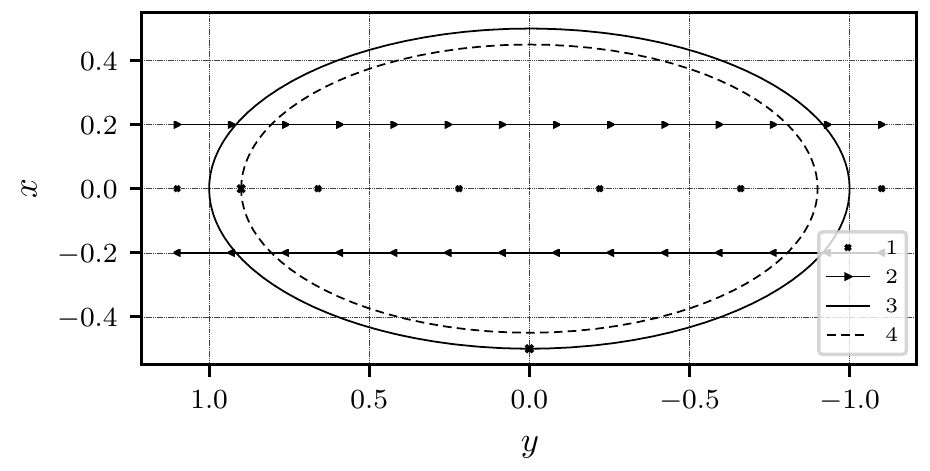}
\caption{Planar Relative Motion Modes for the Clohessy-Wiltshire Problem}
\label{fig:modal_CW}
\end{figure*}

For the planar CW problem, which has LTI dynamics, the decomposition is computed below: 
\begin{equation}
\label{A2d}
[A_{\text{2D}}] = \begin{bmatrix} 0 & 0 & 1 & 0 \\ 0 & 0 & 0 & 1 \\ 3n^{2} & 0 & 0 & 2n \\ 0 & 0 & -2n & 0 \end{bmatrix} = [V][J][V]^{-1}
\end{equation}
\begin{equation}
\label{V_cw2d}
[V] = \begin{bmatrix} 0 & - \frac{2}{3n} & -\frac{1}{2n} & -\frac{1}{2n} \\ 1 & 0 & -\frac{i}{n} & \frac{i}{n} \\ 0 & 0 & -\frac{i}{2} & \frac{i}{2} \\ 0 & 1 & 1 & 1 \end{bmatrix}
\end{equation}
\begin{equation}
\label{J_cw2d}
[J] = \begin{bmatrix} 0 & 1 & 0 & 0 \\ 0 & 0 & 0 & 0 \\ 0 & 0 & ni & 0 \\ 0 & 0 & 0 & -ni \end{bmatrix}
\end{equation}
Using the theory of superposition, the solution to the in-plane dynamics is given below:
\begin{equation}
\label{CWsol1}
\bm{x}(t) = c_{1}\bm{v}_{1}e^{\lambda_{1,2}t} + c_{2}\left(\bm{v}_{1}t + \bm{v}_{2}\right)e^{\lambda_{1,2}t} + c_{3}\bm{v}_{3}e^{\lambda_{3}t} + c_{4}\bm{v}_{4}e^{\lambda_{4}t}
\end{equation}
where $\bm{v}_{i}$ is the $i$th column of $[V]$. Evaluating this at $t = 0$, let the solution constants be defined as $\bm{c} \equiv (c_{1}, \ c_{2}, \ c_{3}, \ c_{4})^{\top}$ Solving $\bm{c} = [V]^{-1}\bm{x}_{0}$ yields the following values for the constants $c_{i}$:
\begin{equation}
\label{CWcs}
\begin{split}
c_{1} = & \ y_{0} - \frac{2}{n}\dot{x}_{0} \\ 
c_{2} = & \ -6nx_{0} - 3\dot{y}_{0} \\
c_{3} = & \ 3nx_{0} + i\dot{x}_{0} + 2\dot{y}_{0} \\ 
c_{4} = & \ 3nx_{0} - i\dot{x}_{0} + 2\dot{y}_{0}
\end{split}
\end{equation}
The solution given by Eq. \eqref{CWsol1} is written in a simpler form, noting $\lambda_{1,2} = 0$, and removing the imaginary part of the constants via the factorization $c_{3} = c_{\mathbb{R}} + ic_{\mathbb{I}}$, and $\bm{v}_{3} = \bm{v}_{\mathbb{R}} + i\bm{v}_{\mathbb{I}}$:
\begin{equation}
\label{CWsol2}
\begin{split}
\bm{x}(t) = & \  c_{1}\bm{v}_{1} + c_{2}\left(\bm{v}_{1}t + \bm{v}_{2}\right) + 2c_{\mathbb{R}}\big(\bm{v}_{\mathbb{R}}\cos{n t} \\ 
& - \bm{v}_{\mathbb{I}}\sin{n t}\big) - 2c_{\mathbb{I}}\left(\bm{v}_{\mathbb{R}}\sin{n t} +  \bm{v}_{\mathbb{I}}\cos{n t}\right)
\end{split}
\end{equation}
Thus, $c_{\mathbb{R}} = 3nx_{0} + 2\dot{y}_{0}$, $c_{\mathbb{I}} = \dot{x}_{0}$, and the fundamental modal solutions weighed by constants $c_{1}$, $c_{2}$, $c_{\mathbb{R}}$, and $c_{\mathbb{I}}$ are plotted in order in Figure \ref{fig:modal_CW}. The initial positions of the oscillatory solutions are marked with an \texttt{x}, and they are scaled such that they don't overlap.

The third and fourth modal solutions are simply two different phases, forming a basis on the 2:1 ellipse, and the first modal solution is a constant offset in the along-track direction, whose scale and direction is determined by the magnitude and sign of $c_{1}$. Comparing this to Eq. \eqref{XandY_CW}, it is clear that the insights of the ROE-based solution have been recovered. For bounded relative motion, the relative orbit is a 2:1 ellipse which can be centered anywhere in the along-track direction. Inspecting the drift solutions in Eq. \eqref{CWsol2} and in Figure \ref{fig:modal_CW}, the magnitude of $c_{2}$ determines the rate of along-track drift, and its sign determines the direction.

As demonstrated by the simple CW example, the strength of a modal decomposition for analysis is that it naturally separates out oscillatory, drifting, and stable/unstable components of the relative motion. For the Keplerian problem, there will always be one relative motion drift mode and no more that three in-plane oscillatory modes. For general periodic orbits, the nature of the relative motion varies based on the dynamics, and concise analytic solutions become impossible. It is in these settings where a modally decomposed solution resolved in favorable coordinates becomes most valuable to the astrodynamicist. However, the application of the theory to the Keplerian case connects strongly to other literature on the topic.

\subsection{Relative Motion Dynamics}
In a coordinate-based dynamics approach (as opposed to the coordinate-invariant approaches of geometric and variational mechanics \cite{lanczos1986}), the choice of coordinates for a system determines the form of the dynamics. For the relative motion problem, the two most common representations are (1) local coordinates of relative position and velocity and (2) the coordinates of orbit element differences. The first is easily physically interpreted, and the latter is often mathematically more convenient. The general perturbed linearized dynamics for both representations are important for this paper and are both reviewed here.

Starting with local coordinates, consider the Cartesian relative state defined previously as $\bm{x} = \left(\bm{\rho}^{\top}, \ \bm{\rho}'^{\top}\right)^{\top}$. Dropping the $c$ subscript for chief orbit parameters, the general linearized dynamics in these coordinates are given below in terms of the chief radial vector $\bm{r}$ and angular velocity vector $\bm{\omega}$ and their derivatives \cite{Casotto}:
\begin{equation}
\label{A_casotto}
\dot{\bm{x}} = \begin{bmatrix} 0_{3\times 3} & I_{3\times 3} \\ \frac{\partial}{\partial \bm{\rho}}\left( \ddot{\bm{r}}_{d} - \ddot{\bm{r}}\right) - \left[\tilde{\dot{\bm{\omega}}}\right] - \left[\tilde{\bm{\omega}}\right]\left[\tilde{\bm{\omega}}\right] & -2\left[\tilde{\bm{\omega}}\right]\end{bmatrix}\bm{x}
\end{equation}
\begin{equation}
\label{omega_cas}
\bm{\omega} = \frac{r}{h}\left(\ddot{\bm{r}}\cdot\hat{\bm{e}}_{n}\right)\hat{\bm{e}}_{r} + \frac{1}{r}\left(\dot{\bm{r}}\cdot\hat{\bm{e}}_{t}\right)\hat{\bm{e}}_{n}
\end{equation}
\begin{equation}
\label{omegadot_cas}
\begin{split}
\dot{\bm{\omega}} = & \ \frac{r}{h}\left(\frac{\dot{r}}{r}\left(\ddot{\bm{r}}\cdot\hat{\bm{e}}_{n}\right) - 2\frac{r}{h}\left(\ddot{\bm{r}}\cdot\hat{\bm{e}}_{t}\right)\left(\ddot{\bm{r}}\cdot\hat{\bm{e}}_{n} \right) + \left(\dddot{\bm{r}}\cdot\hat{\bm{e}}_{n}\right)\right)\hat{\bm{e}}_{r} \\ & + \frac{1}{r}\left(\left(\ddot{\bm{r}}\cdot\hat{\bm{e}}_{t}\right) - 2\frac{\dot{r}}{r}\left(\dot{\bm{r}}\cdot\hat{\bm{e}}_{t}\right)\right)\hat{\bm{e}}_{n}
\end{split}
\end{equation}
where the tilde on a symbol denotes the transformation of its vector into the cross-product matrix, and all matrices appear in square brackets. The above expressions are general, and apply to Keplerian and non-Keplerian dynamics. For the Keplerian case, they simplify significantly into a more common form that can be found in Reference~\citenum{schaub}. An inconvenience of these equations is that they are time-varying if $\dot{\bm{\omega}} \neq \bm{0}$. Further, for the perturbed problem, they can assume a very complicated form and the effects of perturbations are almost irrevocably mixed in with the Keplerian contribution. Furthermore, computation of the plant matrix requires information about the derivative of the force model, shown here explicitly as the jerk, $\dddot{\bm{r}}$. 

An alternative parameterization in terms of orbit element differences $\doe = \oec_{d} - \oec_{c}$ separates out the effect of perturbations from the unperturbed linear dynamics, which are trivial except in one element. This paper will use the differential quasi-nonsingular (QNS) elements given by $\doe = (\delta a, \ \delta\theta, \ \delta i, \ \delta q_{1}, \ \delta q_{2}, \ \delta\Omega)^{\top}$. The Gauss planetary equations are provided below. Note $q_{1} = e\cos{\omega}$, $q_{2} = e\sin{\omega}$, $\theta = \omega+f$ is the argument of latitude, and the other elements are classical semimajor axis $a$, inclination $i$, and right ascension of the ascending node $\Omega$.
\begin{subequations}
\label{GPE_QNS}
\allowdisplaybreaks
\begin{align}
\dot{a} &= \frac{2a^{2}}{h}\left((q_{1}\sin{\theta} - q_{2}\cos{\theta})a_{r} + \frac{p}{r}a_{t}\right)
\\
\dot{\theta} &= \frac{h}{r^{2}} - \frac{r\sin{\theta}\cos{i}}{h\sin{i}}a_{n}
\\
\dot{i} &= \frac{r\cos{\theta}}{h}a_{n}
\\
\dot{q}_{1} &= \frac{p\sin{\theta}}{h}a_{r} + \frac{(p+r)\cos{\theta} + rq_{1}}{h}a_{t} + \frac{r q_{2}\sin{\theta}}{h\tan{i}}a_{n}
\\
\dot{q}_{2} &= -\frac{p\cos{\theta}}{h}a_{r} + \frac{(p+r)\sin{\theta} + rq_{2}}{h}a_{t} - \frac{r q_{1}\sin{\theta}}{h\tan{i}}a_{n}
\\
\dot{\Omega} &= \frac{r\sin{\theta}}{h\sin{i}}a_{n}
\end{align}
\end{subequations}
In the absence of perturbations, the only nonzero term in Eq. \eqref{GPE_QNS} is the true latitude rate $\dot{\theta} = h/r^{2}$. Factoring the Keplerian component of the Jacobian of the right hand side of Eq. \eqref{GPE_QNS} into the mostly zero matrix $[A_{\doe,0}(t)]$, the complicated but typically sub-dominant perturbation-induced component of the Jacobian is written as $[\delta A_{\doe}(t)]$ and the linearized differential QNS dynamics are written concisely below:
\begin{equation}
\label{doe_dyn1}
\delta\dot{\oec} = \Big([A_{\doe,0}(t)] + [\delta A_{\doe}(t)]\Big)\doe
\end{equation}
The benefit of Eq. \eqref{doe_dyn1} over Eq. \eqref{A_casotto} is that the jerk no longer needs to be computed, the effects of perturbations are neatly separated out in the dynamics, and the remaining Keplerian component is fairly simple. A disadvantage is that the differential QNS elements are not as well-suited for geometric interpretation.

References~\citenum{gim1} and \citenum{schaub} discuss the \textit{geometric method}, which relates the relative state in local coordinates to the relative state in differential QNS elements:
\begin{equation}
\label{GeomM1}
\bm{x}(t) = \Big([G_{0}(t)] + [\delta G(t)]\Big)\doe(t) = [G(t)]\doe(t)
\end{equation}
where $[G_{0}(t)]$ captures the Keplerian component of the mapping. The $[\delta G(t)]$ matrix captures the perturbation-induced component of the mapping and is typically sub-dominant to $[G_{0}(t)]$. Reference~\citenum{gim1} demonstrates the derivation of $[\delta G(t)]$ for the $J_{2}$ perturbation. 
\subsection{The Lyapunov-Floquet Transformation in any Coordinates}
The modal decomposition of solutions to a system of ODEs is traditionally defined for autonomous equations, but essentially all non-averaged relative motion except the CW problem is characterized by non-autonomous differential equations. However, the Lyapunov-Floquet transformation \cite{Nayfeh:1979uq} can be used to equate a linear time-varying (LTV) dynamic system with periodic plant matrix to an LTI counterpart via a periodic coordinate transformation:
\begin{equation}
\label{LFintro1}
\bm{x} = [P(t)]\bm{z} = [P(t+T)]\bm{z}
\end{equation}
where $\bm{z}$ represents the coordinate set for the LTI equivalent of the system in $\bm{x}$, with the following simple LTI dynamics:
\begin{equation}
\label{LFintro2}
\dot{\bm{z}} = [\Lambda]\bm{z}
\end{equation}
The LF transformation and the LTI matrix are any pair of matrices $[P(t)]$, $[\Lambda]$ satisfying the following matrix differential equation:
\begin{equation}
\label{LFintro3}
[P(t)]^{-1}\left([A(t)][P(t)] - [\dot{P}(t)]\right) = [\Lambda]
\end{equation}
In analytically solving this equation, which can be challenging, one seeks periodic solutions for the individual elements of $[P(t)]$ while also requiring the elements of $[\Lambda]$ to be constant. In practice, the periodicity conditions for all non-trivial elements of $[P(t)]$ constrain the admissible forms of $[\Lambda]$, but still allow for variations in the values of elements in $[\Lambda]$ depending on the form of $[P(t_{0})]$. As a result, there can be more than a single pair of matrices satisfying Eq. \eqref{LFintro3}. 

A unique definition of the LF transformation is given below using the monodromy matrix. This transformation conveniently equals identity at the epoch time:
\begin{equation}
\label{LFintro4}
[P(t)] = [\Phi(t,t_{0})]e^{-[\Lambda](t - t_{0})}
\end{equation}
\begin{equation}
\label{LFintro5}
[P(t_{0})] = [P(t_{0} + kT)] = [I]
\end{equation}
\begin{equation}
\label{LFintro6}
[\Lambda] = \frac{1}{T}\ln{(\Phi(t_{0}+T,t_{0}))}
\end{equation}
The difficulty of computing the LF transformation varies depending on the coordinates chosen to parameterize the problem. For example, for the Keplerian case, the LF transformation in orbit element differences is shown to be identity except for a single row, whereas the transformation for local coordinates \cite{Sherrill2015LyapunovFloquet}  is much more difficult to identify.

Motivated by the multitude of possible coordinates to parameterize the relative motion problem and the inconvenience of computing the LF transformation from scratch in a given coordinate set, a means to obtain the LF transformation in one set of relative motion coordinates from the transformation in any other set is derived here. 

Let $\bm{x}$ denote the relative state in the desired local coordinates and $\doe$ denote the relative state in the orbit element differences. As already shown with Eq. \eqref{GeomM1}, these two representations are approximately related by an orbit-periodic linear mapping:
\begin{equation}
\label{doeToCart1}
\bm{x} = [G(t)]\doe
\end{equation}
The following linear mapping between the STMs is obtained using Eq. \eqref{doeToCart1}:
\begin{equation}
\label{doeToCart2}
[\Phi_{\bm{x}}(t,t_{0})] =  [G(t)][\Phi_{\doe}(t,t_{0})][G(t_{0})]^{-1}
\end{equation}
The following mapping between the plant matrices can also be shown:
\begin{equation}
\label{doeToCartA}
[A_{\bm{x}}(t)] = [G(t)][A_{\doe}][G(t)]^{-1} + [\dot{G}(t)][G(t)]^{-1}
\end{equation}
Let $[P_{\bm{x}}(t)]$ and $[P_{\doe}(t)]$ denote the LF transformations, transforming the two coordinate sets to their corresponding LTI coordinates:
\begin{equation}
\label{LFtrans_d2C1}
\doe = [P_{\doe}(t)]\bm{z}_{\doe}
\end{equation}
\begin{equation}
\label{LFtrans_d2C2}
\bm{x} = [P_{\bm{x}}(t)]\bm{z}_{\bm{x}}
\end{equation}
These transformations are used to relate the plant matrices for the LTI forms of both coordinates, choosing $[P_{\bm{x}}(t_{0})] = [P_{\doe}(t_{0})] = [I_{6\times6}]$:
\begin{equation}
\label{doeToCart3}
\begin{split}
[\Lambda_{\bm{x}}] = & \ \frac{1}{T}\ln\left([\Phi_{\bm{x}}(t_{0}+T,t_{0})]\right)  \\ = & \ \frac{1}{T}\ln\left([G(t_{0}+T)][\Phi_{\doe}(t_{0}+T,t_{0})][G(t_{0})]^{-1}\right) 
\end{split}
\end{equation}
Noting $[G(t_{0}+T)] = [G(t_{0})]$, the matrix logarithm is factored as follows:
\begin{equation}
\label{doeToCart4}
\begin{split}
[\Lambda_{\bm{x}}] =  & \ [G(t_{0})]\cdot\frac{1}{T}\ln\left([\Phi_{\doe}(t_{0}+T,t_{0})]\right)\cdot[G(t_{0})]^{-1} \\ = & \ [G(t_{0})][\Lambda_{\doe}][G(t_{0})]^{-1}
\end{split}
\end{equation}
The LTI matrix for the local coordinate relative motion representation is simply a change-of-basis of the LTI matrix for the quasi-nonsingular element differences.

Using Eqs. \eqref{doeToCart4} and \eqref{LFintro3}, the following is obtained:
\begin{equation}
\label{transF1}
\begin{split}
& [P_{\bm{x}}]^{-1}\left([A_{\bm{x}}][P_{\bm{x}}] - [\dot{P}_{\bm{x}}]\right) = \\ & [G(t_{0})][P_{\doe}]^{-1}\left([A_{\doe}][P_{\doe}] - [\dot{P}_{\doe}]\right)[G(t_{0})]^{-1}
 \end{split}
\end{equation}
Substituting Eq. \eqref{doeToCartA} and expanding yields
\begin{equation}
\label{transF2}
\begin{split}
[P_{\bm{x}}]^{-1}&[G][A_{\doe}][G]^{-1}[P_{\bm{x}}] \ \\ + & \ [P_{\bm{x}}]^{-1}[\dot{G}][G]^{-1}[P_{\bm{x}}]  - [P_{\bm{x}}]^{-1}[\dot{P}_{\bm{x}}] \\ = & \ [G(t_{0})][P_{\doe}]^{-1}[A_{\doe}][P_{\doe}][G(t_{0})]^{-1} \\ - & \ [G(t_{0})][P_{\doe}]^{-1}[\dot{P}_{\doe}][G(t_{0})]^{-1}
 \end{split}
\end{equation}
This equation is used to show the following relationship between the LF transformations for the two coordinates:
\begin{equation}
\label{transF3}
[P_{\bm{x}}(t)] = [G(t)][P_{\doe}(t)][G(t_{0})]^{-1}
\end{equation}
The LF transformation and LTI form in any set of coordinates can thus be obtained using the corresponding information in another set of coordinates along with the transformation between coordinates via Eqs. \eqref{doeToCart4} and \eqref{transF3}. These relationships hold for linearization about any closed orbit, regardless of whether or not the dynamics are Keplerian.

Via the LF transformation, the modal decomposition of relative motion can be performed in any set of relative state coordinates for any closed orbit:
\begin{equation}
\label{xModalGen}
\bm{x}(t) = \sum_{i=1}^{6}c_{i}[P_{\bm{x}}(t)]\bm{v}_{i}e^{\lambda_{i}(t - t_{0})} 
\end{equation}
where the $\lambda_{i}$ and  $\bm{v}_{i}$ are eigenvalues and eigenvectors of the LTI matrix $[\Lambda_{\bm{x}}]$, and the $c_{i}$ are elements of $\bm{c}$, computed below, noting $[V]$ is constructed column-wise from the $\bm{v}_{i}$:
\begin{equation}
\label{xModalGen2}
\bm{c} = [V]^{-1}\bm{x}(t_{0})
\end{equation}
Eq. \eqref{xModalGen} is of fundamental importance to this paper. Identifying the LF-transformed modal solutions as a desirable set of fundamental solutions, writing $\bm{\xi}_{i}(t) = [P_{\bm{x}}(t)]\bm{v}_{i}e^{\lambda_{i}(t - t_{0})}$ relates Eq. \eqref{xModalGen} back to Eq. \eqref{FS_rm1}. The modal solutions and the LF transformation do not need to be analytically simple, and for general periodic orbits they will not be. It is only desired that they be geometrically as convenient as possible to facilitate relative motion exploration, planning and design. 
All of the developments in this section can be implemented numerically without any great difficulty. Numerical analysis would be the preferred implementation of this approach for perturbed orbits due to the analytical challenges involved. 

Through the modal decomposition using the mapped LF transformation, the astrodynamicist is freed to explore the choice of coordinates that is most desirable for a given application without having to do a prohibitive amount of work when switching coordinates. The only recurring analytic burden is in deriving the necessary linear mapping $[G(t)]$ for any new coordinate representation of interest.

\subsection{The Modal Constants under Perturbations and Control}
It is important to illustrate how the modal methodology can incorporate the effects of perturbations and control. While this paper focuses on geometric solutions to Keplerian motion, this section demonstrates that this theory can be applied to perturbed motion as well.  The application and study of such perturbed solutions is a large body of work that is beyond the scope of this paper.

Let the general modal solution be written in the following compact form from Eq. \eqref{xModalGen}, noting $[\Lambda] = [V][J][V]^{-1}$:
\begin{equation}
\label{VC1}
\bm{x}(t) = [P_{\bm{x}}(t)][V]e^{J(t-t_{0})}\bm{c} = [\Psi(t)] \bm{c}
\end{equation}
For the nominal system, the rate of change of the modal solution satisfies the dynamics based on a linearization about a nominal chief orbit with orbit element history $\oec^{*}$:
\begin{equation}
\label{VC2}
\dot{\bm{x}}(t) = [\dot{\Psi}(t)]\bm{c}  = [A(\oec^{*},t)]\bm{x}
\end{equation}

To model the effect of perturbations, the constants $\bm{c}$ are allowed to vary such that the state rate of the modal solution matches the perturbed dynamics, which are given below: 
\begin{equation}
\label{xNLD1}
\dot{\bm{x}} = \bm{f}(\bm{x},\bm{u},t)
\end{equation}

For the perturbed problem, the formerly constant quantities in $\bm{c}$ are made to vary in a way such that Eq. \eqref{VC1} still accurately describes the motion of $\bm{x}(t)$, where $[V]$, $[J]$, and $[P_{\bm{x}}(t)]$ still track the nominal chief orbit from which they were computed. Differentiating Eq. \eqref{VC1}, this requirement is written mathematically:
\begin{equation}
\label{VCr3}
\begin{split}
 \dot{\bm{x}}(t) = & \ \frac{\text{d}}{\text{d}t}\left([P_{\bm{x}}(t)][V]e^{J(t-t_{0})}\bm{c}(t)\right) \\ = & \ \frac{\partial\bm{x}}{\partial t} + \frac{\partial\bm{x}}{\partial\bm{c}}\dot{\bm{c}} = \bm{f}(\bm{x},\bm{u},t)
\end{split}
\end{equation}
In consistency with Eq. \eqref{VC2}, the following osculating condition needs to be achieved at all times:
\begin{equation}
\label{VCr5}
\frac{\partial\bm{x}}{\partial t} = [A(\oec^{*},t)]\bm{x}
\end{equation}
The following dynamics satisfy the osculating condition:
\begin{equation}
\label{VC6}
\begin{split}
\dot{\bm{c}} = & \ \left(\frac{\partial\bm{x}}{\partial\bm{c}}\right)^{-1}\left( \bm{f}(\bm{x},\bm{u},t) - [A(\oec^{*},t)]\bm{x}\right) \\ = & \ [\Psi(t)]^{-1}\left( \bm{f}(\bm{x},\bm{u},t) - [A(\oec^{*},t)]\bm{x}\right)
\end{split}
\end{equation}

From Eq. \eqref{VC6}, the effects of perturbations, nonlinearities, and control on $\bm{c}$ can be easily explored. In the case that the linearization about the nominal chief orbit used to compute the modes is accurate, the controlled linearized dynamics of $\bm{c}$ reduce to the following:
\begin{equation}
\label{VC6Sub1Simple}
\dot{\bm{c}} = [\Psi(t)]^{-1}[B_{\bm{x}}]\bm{u}
\end{equation}
where $[B_{\bm{x}}] = [0_{3\times3} \ I_{3\times3}]^{\top}$ if $\bm{x}$ is in Cartesian coordinates. 

\section{Application to the Keplerian Relative Motion Problem}
In this section, the LF transformation is obtained for Keplerian dynamics of any eccentricity in QNS element differences, and this is analytically transformed to LF transformations in Cartesian and spherical coordinates. It is shown that the modal solutions in Cartesian and spherical coordinates are different. This is an interesting result that illustrates how the choice of working coordinates can affect the complexity of the modal solutions.
\subsection{Orbit Element Differences}
For two-body dynamics, the relative motion dynamics in QNS elements can be shown to take the following simplified form by transforming the independent variable from $t$ to $\theta$:
\begin{equation}
\label{QNSeom1}
\doe' = \begin{bmatrix} 0 & 0 & 0 & 0 & 0 & 0 \\ -\frac{3}{2a} & \frac{2(q_{2}\text{c}{\theta} - q_{1}\text{s}{\theta})}{\kappa} & 0 & \frac{3q_{1}}{\eta^{2}} + \frac{2\text{c}{\theta}}{\kappa} &  \frac{3q_{2}}{\eta^{2}} + \frac{2\text{s}{\theta}}{\kappa} & 0 \\ 0 & 0 & 0 & 0 & 0 & 0 \\ 0 & 0 & 0 & 0 & 0 & 0 \\ 0 & 0 & 0 & 0 & 0 & 0 \\ 0 & 0 & 0 & 0 & 0 & 0  \end{bmatrix}\doe
\end{equation}
where $\text{s}=\sin{( \ )}$, $\text{c}=\cos{( \ )}$ and the shorthand quantities $\eta$, $\kappa$ and $\kappa_{0}$ are defined below. 
\begin{equation}
\label{eta0}
 \eta = \sqrt{1 - q_{1}^{2} - q_{2}^{2}}
 \end{equation}
\begin{equation}
\label{kappa}
\kappa = 1 + q_{1}\cos{\theta} + q_{2}\sin{\theta}
\end{equation}
\begin{equation}
\label{kappa0}
\kappa_{0} = 1 + q_{1}\cos{\theta_{0}} + q_{2}\sin{\theta_{0}}
\end{equation}
The dynamics are $\doe' = \dfrac{\text{d}}{\text{d}\theta}\left(\doe\right) = \dfrac{1}{\dot{\theta}}\delta\dot{\oec}$, so the plant matrix in Eq. \eqref{QNSeom1} is $[\tilde{A}(\theta)] = \frac{1}{\dot{\theta}}[A(\theta)]$.

A Lyapunov-Floquet transformation of Eq. \eqref{QNSeom1} is sought, because $[\tilde{A}(\theta)] = [\tilde{A}(\theta + 2\pi)]$. To differentiate the LTI system for this new choice of independent variable, let the LTI coordinates $\bm{\chi}$ be used instead of $\bm{z}$ when $\theta$ is the independent variable, with associated LTI plant matrix $[R]$ instead of $[\Lambda]$, and LF transformation $[P(\theta)]$:
\begin{equation}
\label{QNStrans1}
\doe = [P(\theta)]\bm{\chi}
\end{equation}
\begin{equation}
\label{QNS_LTItheta}
\bm{\chi}' = [R]\bm{\chi}
\end{equation}
The LF transformation $[P(\theta)]$ solves an equivalent of Eq. \eqref{LFintro3}:
\begin{equation}
\label{QNSeom2}
[P(\theta)]^{-1}\left([\tilde{A}(\theta)][P(\theta)] - [P'(\theta)]\right) = [R]
\end{equation}
The LF transformation sought is determined to have the following simple form:
\begin{equation}
\label{PofTheta_QNS1}
[P(\theta)] =  \begin{bmatrix} 1 & 0 & 0 & 0 & 0 & 0 \\ P_{21}(\theta) & P_{22}(\theta) & 0 & P_{24}(\theta) & P_{25}(\theta) & 0 \\ 0 & 0 & 1 & 0 & 0 & 0 \\ 0 & 0 & 0 & 1 & 0 & 0 \\ 0 & 0 & 0 & 0 & 1 & 0 \\ 0 & 0 & 0 & 0 & 0 & 1 \end{bmatrix}
\end{equation}
This reduces the number of scalar differential equations in Eq. \eqref{QNSeom2} to four:
\begin{subequations}
\label{PrDEList}
\begin{align}
\tilde{A}_{21} + \tilde{A}_{22}(\theta)P_{21} - P'_{21} = & \ R_{21}P_{22}(\theta)
\\
\tilde{A}_{22}(\theta)P_{22} - P'_{22} = & \ R_{22}P_{22}(\theta)
\\
\tilde{A}_{24}(\theta) + \tilde{A}_{22}(\theta)P_{24} - P'_{24} = & \ R_{24}P_{22}(\theta)
\\
\tilde{A}_{25}(\theta) + \tilde{A}_{22}(\theta)P_{25} - P'_{25} = & \ R_{25}P_{22}(\theta)
\end{align}
\end{subequations}
These equations are solved, starting with $P_{22}(\theta)$, and enforcing a periodicity condition for each solution. This is demonstrated only for $P_{22}(\theta)$, whose general solution is given below:
\begin{equation}
\label{P22Theta1}
P_{22}(\theta) = c_{1}\left(1 + q_{1}\cos{\theta} + q_{2}\sin{\theta}\right)^{2}e^{-R_{22}\theta}
\end{equation}
where $c_{1}$ is an integration constant. The periodicity condition $P_{22}(\theta) = P_{22}(\theta + 2\pi)$ yields $R_{22} = 0$, and the resulting form for $P_{22}(\theta)$ is substituted into the other differential equations, which are solved for their own periodic solutions. An additional constraint is that $[P(\theta_{0})] = [I]$ to obtain the desired LF transformation discussed in Section 2.3. The finalized nonzero components of the LF transformation are given below, along with the LTI matrix:
\begin{subequations}
\label{Pfinal2}
\allowdisplaybreaks
\begin{align}
\begin{split}
P_{21}(\theta) = & \  \frac{\kappa^{2}}{2a}\left(\mathbb{F}_{21}(\theta_{0}) - \mathbb{F}_{21}(\theta)\right) 
\end{split} \\
\begin{split}
\mathbb{F}_{21}(\theta) = & \ \frac{6}{\eta^{3}}\left(\tan^{-1}{\left(\frac{q_{2} + (1 - q_{1})\tan{\left(\frac{\theta}{2}\right)}}{\sqrt{1 - q_{1}^{2} - q_{2}^{2}}}\right)} -\frac{\theta}{2}\right) \\ + & \ \frac{3\left(q_{2} + (q_{1}^{2} + q_{2}^{2})\sin{\theta}\right)}{q_{1}(q_{1}^{2} + q_{2}^{2} - 1)\kappa}
\end{split} \\
\begin{split}
P_{22}(\theta) = & \ \frac{\kappa^{2}}{\kappa_{0}^{2}}
\end{split} \\ 
\begin{split}
P_{24}(\theta) = & \ \frac{\kappa^{2}}{4\left(q_{1}^{2} + q_{2}^{2} - 1\right)}\left(\mathbb{F}_{24}(\theta_{0}) - \mathbb{F}_{24}(\theta)\right)
\end{split} \\
\begin{split}
\mathbb{F}_{24}(\theta) = & \ \frac{4(q_{2} + \sin{\theta})}{\kappa^{2}} + \frac{4\sin{\theta}}{\kappa}
\end{split} \\
\begin{split}
P_{25}(\theta) = & \ \frac{\kappa^{2}}{4\left(q_{1}^{2} + q_{2}^{2} - 1\right)}\left(\mathbb{F}_{25}(\theta_{0}) - \mathbb{F}_{25}(\theta)\right) 
\end{split} \\
\begin{split}
\mathbb{F}_{25}(\theta) = & \ \frac{4(1 - q_{1}^{2} + q_{2}\sin{\theta})}{q_{1}\kappa^{2}}  + \frac{4q_{2}\sin{\theta}}{q_{1}\kappa}
\end{split} 
\end{align}
\end{subequations}
\begin{equation}
\label{RmatFinal}
[R] = \begin{bmatrix} 0 & 0 & 0 & 0 & 0 & 0 \\ -\frac{3a\eta}{2r_{0}^{2}} & 0 & 0 & 0 & 0 & 0\\ 0 & 0 & 0 & 0 & 0 & 0 \\ 0 & 0 & 0 & 0 & 0 & 0 \\ 0 & 0 & 0 & 0 & 0 & 0 \\ 0 & 0 & 0 & 0 & 0 & 0 \end{bmatrix}
\end{equation}
Note that there is a singularity in Eq. \eqref{Pfinal2} for $P_{25}$ for the case of $q_{1} = 0$. 

The Lyapunov-Floquet transformation and LTI form for the case that $t$ is the independent variable instead of $\theta$ is now discussed. Due to the explicit appearance of the intermediate variable $\theta$, this alternate form offers no computational advantages. It is however slightly simpler. First, the nonzero element of the new LTI matrix $[\Lambda_{\doe}]$ is $nR_{21}$:
\begin{equation}
\label{Gamma21simple}
\Lambda_{21} = -\frac{3a\eta}{2r_{0}^{2}}n
\end{equation}
where $n = \sqrt{\mu/a^{3}}$ is the mean motion. In this case, the Lyapunov-Floquet transformation takes on a slightly simpler form, with the $P_{21}(\theta)$ term in Eq. \eqref{PofTheta_QNS1} reducing to zero, and all other components unaffected. With this modified transformation, the equation for $\delta\theta$ reduces to a familiar form:
\begin{equation}
\label{dtheta_sol2}
\delta\theta = P_{22}(\theta)R_{21}n(t-t_{0})\delta a + P_{22}(\theta)\delta\theta_{0} + P_{24}(\theta)\delta q_{1} + P_{25}(\theta)\delta q_{2}
\end{equation}
This expression is analytically equivalent to its counterpart in Eq. (14.129) of Reference \citenum{schaub}, though derived by quite a different process. 
Exploiting the equivalence of Eq. \eqref{dtheta_sol2} to Eq. (14.129) in Reference  \citenum{schaub}, alternate expressions can be obtained for $P_{24}(\theta)$ and $P_{25}(\theta)$ from Eq. (14.130). The alternate expression for $P_{25}(\theta)$ is notably nonsingular for $q_{1} = 0$.

The mapping of LF transformations is to be applied for two alternate sets of relative motion coordinates. For this, Eqs. \eqref{doeToCart4} and \eqref{transF3} are repeated with $\theta$ instead of $t$:
\begin{equation}
\label{LTI_map1}
[R_{\bm{x}}] = [G(\theta_{0})][R_{\doe}][G(\theta_{0})]^{-1}
\end{equation}
\begin{equation}
\label{LF_map1}
[P_{\bm{x}}(\theta)] = [G(\theta)][P_{\doe}(\theta)][G(\theta_{0})]^{-1}
\end{equation}
where $[R_{\doe}]$ is given by Eq. \eqref{RmatFinal} and $[P_{\doe}(\theta)]$ is given by Eqs. \eqref{PofTheta_QNS1} and \eqref{Pfinal2}.

\subsection{Cartesian Coordinates}
Let $\bm{x}_{c} = (x, \ y, \ z, \ \dot{x}, \ \dot{y}, \ \dot{z})^{\top}$ denote the state in local Cartesian coordinates. For these coordinates, the linearized coordinate transformation $[G(\theta)]$ from QNS orbit element differences is reproduced below \cite{schaub}:
\small
\begin{multline}
\label{P_linear}
[G_{\bm{x}_{c}}]  = \left[
  \begin{matrix}
     \frac{r}{a} & \frac{v_{r}}{v_{t}}r & 0 \\ 
     0 & r & 0 \\
    0 & 0 & r\text{s}{\theta} \\
    -\frac{v_{r}}{2a} & \left(\frac{1}{r} - \frac{1}{p}\right)h & 0 \\
    -\frac{3v_{t}}{2a} & -v_{r} & 0 \\
    0 & 0 & (v_{t}\text{c}{\theta} + v_{r}\text{s}{\theta}) 
  \end{matrix}\right.                
  \left.
  \begin{matrix}
    -\frac{r}{p}(2aq_{1} + r\text{c}{\theta}) & -\frac{r}{p}(2aq_{2} + r\text{s}{\theta}) & 0 \\
    0 & 0 & r\text{c}{i} \\
    0 & 0 & -r\text{c}{\theta}\text{s}{i} \\
   \frac{1}{p}(v_{r}aq_{1} + h\text{s}{\theta}) & \frac{1}{p}(v_{r}aq_{2} - h\text{c}{\theta}) & 0 \\
    \frac{1}{p}(3v_{t}aq_{1} + 2h\text{c}{\theta}) & \frac{1}{p}(3v_{t}aq_{2} + 2h\text{s}{\theta}) & v_{r}\text{c}{i} \\
   0 & 0 & (v_{t}\text{s}{\theta} - v_{r}\text{c}{\theta})\text{s}{i}
  \end{matrix}\right]
\end{multline}
\normalsize
where $v_{r} = \dot{r}$ and $v_{t} = r\dot{\theta}$, and the shorthand $\text{s}$ and $\text{c}$ are sine and cosine.
For the inverse of Eq. \eqref{P_linear}, see Reference \citenum{schaub}. 

Solving Eq. \eqref{LTI_map1}, the Cartesian LTI matrix $[R_{\bm{x}_{c}}]$ is obtained, which can be expressed in a highly compact form:
\begin{equation}
\label{RmatCart}
[R_{\bm{x}_{c}}] = \frac{2R_{21}a}{\gamma}
  \begin{bmatrix}
    A(B+2) & A^{2} & 0 & A^{2}C &  -A(B+1)C  & 0 \\
    -(B+1)(B+ 2) & -A(B+1) & 0 & -A(B+1)C & (B+1)^{2}C & 0 \\
    0 & 0 & 0 & 0 & 0 & 0 \\
    B(B+2)/C & AB/C & 0 & AB & -B(B+1) & 0 \\
    A(B+2)/C & A^{2}/C & 0 & A^{2} & -A(B+1) & 0 \\
    0 & 0 & 0 & 0 & 0 & 0
  \end{bmatrix}
\end{equation}
where $R_{21}$ is the nonzero $(2, 1)$ element of $[R_{\doe}]$ in Eq. \eqref{RmatFinal} and the shorthand quantities $\gamma$, $A$, $B$, $C$ are defined below:
\begin{equation}
\label{sh1}
\gamma = q_{1}^ {2} + q_{2}^{2} - 1 = A^{2} + B^{2} - 1
\end{equation}
\begin{equation}
\label{sh2}
A = -\frac{v_{r,0}p}{v_{t,0}r_{0}} = q_{2}\cos{\theta_{0}} - q_{1}\sin{\theta_{0}}
\end{equation}
\begin{equation}
\label{sh3}
B = \frac{p}{r_{0}} - 1 = q_{1}\cos{\theta_{0}} + q_{2}\sin{\theta_{0}}
\end{equation}
\begin{equation}
\label{sh4}
C = \frac{hr_{0}^{2}}{a\mu\gamma}
\end{equation}
The true and generalized eigenvectors of $[R_{\bm{x}_{c}}]$ are given as the columns of $[V_{R_{\bm{x}_{c}}}]$ below. The matrix $[R_{\bm{x}_{c}}]$ has six zero eigenvalues, with geometric multiplicity 5. 
\begin{equation}
\label{VforRmatCart}
[V_{R_{\bm{x}_{c}}}] = \begin{bmatrix} 0 & 0 & 0 & 0 & -\frac{2R_{21}a}{\gamma}A(B+1)C & 0 \\
1 & 0 & 0 & 0 & \frac{2R_{21}a}{\gamma}(B+1)^{2}C & 0 \\
0 & 1 & 0 & 0 & 0 & 0 \\
-\frac{1}{C} & 0 & 1 & 0 & -\frac{2R_{21}a}{\gamma}B(B+1) & 0 \\
0 & 0 & \frac{A}{B+1} & 0 & -\frac{2R_{21}a}{\gamma}A(B+1) & 1 \\
0 & 0 & 0 & 1 & 0 & 0
\end{bmatrix}
\end{equation}

Note that both $C$ and the scaling term on the fifth column of $[V_{R_{\bm{x}}}]$ can be expressed in terms of $A$ and $B$:
\begin{equation}
\label{CfromAB}
C = -\frac{(1 - A^{2} - B^{2})^{3/2}}{(B+1)^{2}n}
\end{equation}
\begin{equation}
\label{scaleABC}
\frac{2R_{21}a}{\gamma} = \frac{3(B+1)^{2}}{(1-A^{2}-B^{2})^{5/2}}
\end{equation}
Because typically $|C| \gg 1$, the scaling of the fifth column of $[V_{R_{\bm{x}}}]$ can be much larger than the others. 

The general solution of the LTI form for the Cartesian coordinates is given below in terms of the columns of $[V_{R_{\bm{x}_{c}}}]$ and the solution constants, to be defined shortly:
\begin{equation}
\label{xLTIsol}
\bm{\chi}_{\bm{x}_{c}}(\theta) = \sum_{i=1}^{5}c_{i}\bm{v}_{i} + c_{6}\left(\bm{v}_{5}(\theta-\theta_{0}) + \bm{v}_{6}\right)
\end{equation}

The LF transformation for Cartesian coordinates maps the solution given by Eq. \eqref{xLTIsol} back to Cartesian coordinates via $\bm{x}_{c} = [P_{\bm{x}_{c}}]\bm{\chi}_{\bm{x}_{c}}$. It is computed using the mapping from Eq. \eqref{Pfinal2} given by Eq. \eqref{LF_map1}, making use of Eq. \eqref{P_linear}. This is significantly easier than solving differential equations for its elements. The resulting LF transformation is a product of analytic matrices, and can be evaluated efficiently. 

Using the inverse of Eq. \eqref{VforRmatCart}, the constant vector $\bm{c} = (c_{1}, \ c_{2}, \ c_{3}, \ c_{4}, \ c_{5}, \ c_{6})^{\top}$ is given by $\bm{c} = [V_{R_{\bm{x}_{c}}}]^{-1}\bm{\chi}_{\bm{x}_{c}}(\theta_{0})$:
\begin{subequations}
\label{c123}
\allowdisplaybreaks
\begin{align}
c_{1} = & \ -\frac{v_{t,0}}{v_{r,0}}x_{0} + y_{0} \\
c_{2} = & \ z_{0} \\
c_{3} = & \ \frac{1}{C}\left(-\frac{v_{t,0}r_{0}}{v_{r,0}p}x_{0} + y_{0} + C\dot{x}_{0}\right) \\
c_{4} = & \ \dot{z}_{0} \\
c_{5} = & \  -\frac{(1 - e^{2})v_{t,0}}{3v_{r,0}}n\left(\frac{r_{0}}{p}\right)^{2}x_{0} \\
c_{6} = & \ \frac{\left(\frac{p}{r_{0}}+1\right)\frac{p}{r_{0}}n}{(1 - e^{2})^{3/2}}x_{0} + \frac{1}{C}\frac{v_{r,0}}{v_{t,0}}y_{0} + \frac{v_{r,0}}{v_{t,0}}\dot{x}_{0} + \dot{y}_{0}
\end{align}
\end{subequations}
The expression for $c_{6}$ reduces to the Clohessy-Wiltshire no-drift constraint $c_{6} = 2nx_{0} + \dot{y}_{0}$ when $e = 0$, so $c_{6}$ captures the degree of drift. It is better understood as a linearized measure of $\delta a$. In particular, if $\delta a = 0$, this quantity should be zero as well. The terms $c_{1}$, $c_{3}$, and $c_{5}$ are affiliated with the in-plane modes, and $c_{2}$ and $c_{4}$ are associated with the two out-of-plane oscillatory modes.

The analytic perspective offered by Eq. \eqref{c123} is very useful. First, the out-of-plane motion is decoupled from the in-plane motion. Additionally, none of the in-plane constants except $c_{6}$ are functions of $\dot{y}_{0}$. In the case that the degree of drift is specified via a fixed value of $c_{6}$, a select initial in-plane component of the position $(x_{0}, \ y_{0})$ forms a point of intersection of all possible in-plane relative motions in a one-parameter variation, based on the value of $\dot{x}_{0}$. The constants $c_{1}$ and $c_{5}$ are fixed by the choice of initial position, and only the value of $c_{3}$ varies as the value of $\dot{x}_{0}$ is varied. Additionally, only two of the in-plane mode constants, $c_{3}$ and $c_{6}$, can be changed with a single impulsive maneuver. The constants $c_{1}$ and $c_{5}$ can only be changed with a two-burn sequence. 

On the topic of maneuvers and the drift constant, an additional result can be determined from the constant $c_{6}$. For single-maneuver changes to bounded relative motion, for which $c_{6} = 0$, the in-plane component of the thruster direction is constrained to a line:
\begin{equation}
\label{c6toVc}
\Delta v_{y} = - \frac{v_{r,0}}{v_{t,0}}\Delta v_{x}
\end{equation}
Any maneuver not satisfying this constraint will introduce drift to the relative motion. For two-burn maneuvers, the orbit must be parameterized in terms of two sets of constants $\bm{c}$ and $\bm{c}'$ at the two distinct maneuver points in the orbit. Note that the following equation can be used to map constants $\bm{c}$ for a choice of epoch anomaly $\theta_{0}$ to a new epoch angle $\theta_{0}'$:
\begin{equation}
\label{cMap}
\bm{c}(\theta_{0}') = [V(\theta_{0}')]^{-1}[\Phi_{\bm{x}_{c}}(\theta_{0}',\theta_{0})][V(\theta_{0})]\bm{c}(\theta_{0})
\end{equation}

The inverse of Eq. \eqref{VforRmatCart} becomes singular when $A = 0$. This is equivalent to whenever $e\sin{f_{0}} = 0$, or whenever $e$ and/or $f_{0}$ is equal to zero. However, the issue can be remedied by evaluating the expression with the offending terms set to a small number $\epsilon$ instead of exactly zero. 
For orbits of nonzero eccentricity, the singularity issue can also always be avoided by selecting $f_{0} \neq k\pi$ for integers $k$. 

For Keplerian orbits, the general linear relative motion problem in Cartesian coordinates are studied in terms of individual modes via the following:
\begin{equation}
\label{xModalSol}
\bm{x}_{c}(\theta) = \sum_{i=1}^{5}c_{i}[P_{\bm{x}_{c}}(\theta)]\bm{v}_{i} + c_{6}[P_{\bm{x}_{c}}(\theta)]\left(\bm{v}_{5}(\theta-\theta_{0}) + \bm{v}_{6}\right)
\end{equation}
where the transformation $[P_{\bm{x}_{c}}]$ given by Eqs. \eqref{LF_map1} and \eqref{P_linear} is required to evaluate this expression and the $\bm{v}_{i}$ are the columns of Eq.  \eqref{VforRmatCart}. Compare Eq. \eqref{xModalSol} to Eqs. \eqref{FS_rm1} and \eqref{xModalGen}.
The individual modal solutions for the Cartesian modal decomposition are plotted and studied in Section 3.4.

\subsection{Spherical Coordinates}
The local spherical coordinate representation is given by $\bm{x}_{s} = \left(\delta r, \ \theta_{r}, \ \phi_{r}, \ \delta\dot{r}, \ \dot{\theta}_{r}, \ \dot{\phi}_{r}\right)^{\top}$. It has the advantage over the local Cartesian coordinate representation of better capturing the curvature characteristic of large along-track separations and large out-of-plane motion. This makes it a more accurate representation for relative motion problems with large along-track separations.

The relative state in local spherical coordinates is obtained from local Cartesian coordinates as below: 
\begin{subequations}
\label{CartToSph}
\allowdisplaybreaks
\begin{align}
\delta r = & \sqrt{(r_{c} + x)^{2} + y^{2} + z^{2}} - r_{c} \\
\theta_{r} = & \tan^{-1}{\left(\frac{y}{r_{c} + x}\right)} \\
\phi_{r} = & \sin^{-1}{\left(\frac{z}{\sqrt{(r_{c} + x)^{2} + y^{2} + z^{2}}}\right)} \\
\delta\dot{r} = & \frac{(r_{c} + x)(\dot{r}_{c} + \dot{x}) + y\dot{y} + z\dot{z}}{\sqrt{(r_{c} + x)^{2} + y^{2} + z^{2}}} - \dot{r}_{c} \\
\dot{\theta}_{r} = & \ \frac{(r_{c} + x)\dot{y} - y(\dot{r}_{c} + \dot{x})}{(r_{c} + x)^{2} + y^{2}} \\
\dot{\phi}_{r} = & \frac{(r_{c} + \delta r)\dot{z} - (\dot{r}_{c} + \delta\dot{r})z}{(r_{c}+\delta r)^{2}\sqrt{1 - \frac{z^{2}}{(r_{c} + \delta r)^{2}}}}
\end{align}
\end{subequations}

The position components of the inverse transformation are given below, for which the corresponding velocities can be obtained by differentiation:
\begin{subequations}
\label{SphToCart}
\begin{align}
x = & \ (r_{c} + \delta r)\cos{\theta_{r}}\cos{\phi_{r}} - r_{c} \\
y = & \ (r_{c} + \delta r)\sin{\theta_{r}}\cos{\phi_{r}} \\
z = & \ (r_{c} + \delta r)\sin{\phi_{r}}
\end{align}
\end{subequations}
The linearized transformation between the local spherical and Cartesian coordinate representations is given below:
\begin{subequations}
\label{CartToSphLin}
\begin{align}
\delta r \approx & \ x \\
\theta_{r} \approx & \ y/r_{c} \\
\phi_{r} \approx & \ z/r_{c} \\
\delta\dot{r} \approx & \ \dot{x} \\
\dot{\theta}_{r} \approx & \ \dot{y}/r_{c} - (\dot{r}_{c}/r_{c}^{2})y \\
\dot{\phi}_{r} \approx & \ \dot{z}/r_{c} - (\dot{r}_{c}/r_{c}^{2})z
\end{align}
\end{subequations}
The nonlinear transformation from orbit element differences to local spherical coordinates is discussed in Reference \citenum{Han2019OEstoSph}. For this work, the linearized transformation is derived for small angles $\theta_{r}$ and $\phi_{r}$. 
The result is given below:
\begin{equation}
\label{P_linear_sph}
[G_{\bm{x}_{s}}] =   \left[
  \begin{matrix}
     \frac{r}{a} & \frac{v_{r}}{v_{t}}r & 0 \\ 
    0 & 1 & 0  \\
    0 & 0 & \text{s}{\theta}  \\
    -\frac{v_{r}}{2a} & \left(\frac{1}{r} - \frac{1}{p}\right)h & 0 \\
    -\frac{3\dot{\theta}}{2a} & -2\frac{v_{r}}{r} & 0 \\
    0 & 0 & \dot{\theta}\text{c}{\theta} 
  \end{matrix}\right.                
  \left.
  \begin{matrix}
   -\frac{r}{p}(2aq_{1} + r\text{c}{\theta}) & -\frac{r}{p}(2aq_{2} + r\text{s}{\theta}) & 0 \\
   0 & 0 & \text{c}{i}  \\
    0 & 0 & -\text{c}{\theta}\text{s}{i} \\
   \frac{1}{p}(v_{r}aq_{1} + h\text{s}{\theta}) & \frac{1}{p}(v_{r}aq_{2} - h\text{c}{\theta}) & 0 \\
   \frac{\dot{\theta}}{p}(3aq_{1} + 2r\text{c}{\theta}) & \frac{\dot{\theta}}{p}(3aq_{2} + 2r\text{s}{\theta}) & 0 \\
   0 & 0 & \dot{\theta}\text{s}{\theta}\text{s}{i}
  \end{matrix}\right]
\end{equation}
Note the similarity of Eqs. \eqref{P_linear} and \eqref{P_linear_sph}. The first and fourth rows are identical.

Solving Eq. \eqref{LTI_map1}, the spherical coordinate LTI matrix $[R_{\bm{x}_{s}}]$ is obtained, expressed below in a form similar to Eq. \eqref{RmatCart}:
\begin{equation}
\label{RmatSph}
[R_{\bm{x}_{s}}] = \frac{2R_{21}a}{\gamma}\left[
  \begin{matrix}
    A(B+2) & 0 & 0 \\ 
    \frac{(B+1)^{2}(B+2)}{\gamma a} & 0 & 0  \\
    0 & 0 & 0  \\
   \frac{B(B+2)}{C} & 0 & 0 \\
    -\frac{2A(B+1)(B+2)}{\gamma a C} & 0 & 0 \\
    0 & 0 & 0 
  \end{matrix}\right.                
  \left.
  \begin{matrix}
   A^{2}C & \gamma a AC  & 0 \\
   \frac{AC(B+1)^{2}}{\gamma a} & (B+1)^{2}C & 0  \\
    0 & 0 & 0\\
   AB & \gamma a B & 0 \\
   -\frac{2A^{2}(B+1)}{\gamma a} & -2A(B+1) & 0 \\
   0 & 0 & 0
  \end{matrix}\right]
\end{equation}
The true and generalized eigenvectors of $[R_{\bm{x}_{s}}]$ are given as the columns of $[V_{R_{\bm{x}_{s}}}]$ below. Like the LTI matrix for QNS element differences and Cartesian coordinates, the matrix $[R_{\bm{x}_{s}}]$ has six zero eigenvalues, with geometric multiplicity 5. 
\begin{equation}
\label{VforRmatSph}
[V_{R_{\bm{x}_{s}}}] = \begin{bmatrix} 0 & 0 & 0 & 0 & \frac{2R_{21}a}{\gamma}AC\gamma a & 0 \\
1 & 0 & 0 & 0 & \frac{2R_{21}a}{\gamma}(B+1)^{2}C & 0 \\
0 & 1 & 0 & 0 & 0 & 0 \\
0 & 0 & 1 & 0 & \frac{2R_{21}a}{\gamma}B\gamma a & 0 \\
0 & 0 & -\frac{A}{\gamma a} & 0 & -\frac{4R_{21}a}{\gamma}A(B+1) & 1 \\
0 & 0 & 0 & 1 & 0 & 0
\end{bmatrix}
\end{equation}
Analogously as for Cartesian coordinates, the general solution of the LTI form for the spherical coordinates is given below in terms of the columns of $[V_{R_{\bm{x}_{s}}}]$. The constant vector $\bm{c} = (c_{1}, \ c_{2}, \ c_{3}, \ c_{4}, \ c_{5}, \ c_{6})^{\top}$ is given by $\bm{c} = [V_{R_{\bm{x}_{s}}}]^{-1}\bm{\chi}_{\bm{x}_{s}}(\theta_{0})$:
\begin{subequations}
\label{c123sph}
\begin{align}
c_{1} = & \ -\frac{v_{t,0}}{v_{r,0}r}\delta r_{0} + \theta_{r,0} \\
c_{2} = & \ \phi_{r,0} \\
c_{3} = & \ \frac{1}{C}\left(\frac{\left(1 - \frac{r_{0}}{p}\right)v_{t,0}}{v_{r,0}}\delta r_{0} + C\delta \dot{r}_{0} \right) \\
c_{4} = & \ \dot{\phi}_{r,0} \\
c_{5} = & \  -\frac{v_{t,0}}{3v_{r,0}a}n\left(\frac{r_{0}}{p}\right)\delta r_{0} \\
c_{6} = & \ \frac{\mu}{hr_{0}^{2}}\left(1 + \frac{p}{r_{0}}\right)\delta r_{0} + \frac{v_{r,0}}{v_{t,0}r_{0}}\delta\dot{r}_{0} + \dot{\theta}_{r,0} 
\end{align}
\end{subequations}
Note that the equation for $c_{6}$ in Eq. \eqref{c123sph} is zero when $\delta a = 0$. It represents a more concise local coordinate no-drift condition than its counterpart in Eq. \eqref{c123}.

For Keplerian orbits, the general linear relative motion problem in spherical coordinates are studied in terms of individual modes via the following:
\begin{equation}
\label{xsphModalSol}
\bm{x}_{s}(\theta) = \sum_{i=1}^{5}c_{i}[P_{\bm{x}_{s}}(\theta)]\bm{v}_{i} + c_{6}[P_{\bm{x}_{s}}(\theta)]\left(\bm{v}_{5}(\theta-\theta_{0}) + \bm{v}_{6}\right)
\end{equation}
where the transformation $[P_{\bm{x}_{s}}]$ given by Eq. \eqref{LF_map1} is required to evaluate this expression and the $\bm{v}_{i}$ are the columns of Eq.  \eqref{VforRmatSph}. Note that the singularity properties of the Cartesian and spherical coordinate representations are the same.  

To project the spherical coordinate results into Cartesian coordinates, there are two options. The nonlinear transformation given by Eq. \eqref{SphToCart} and its first derivative can be used, or the inverse of the linearized transformation given by Eq. \eqref{CartToSphLin} can be used. The former is a more accurate transformation that will capture the curvature of the relative motion missed by the Cartesian representation, while the latter transformation has the benefit of being linear, but lacks the additional accuracy offered by the nonlinear transformation. The results of the linearized Cartesian and spherical coordinate representations can be made completely equivalent via the linearized transformation Eq. \eqref{CartToSphLin} and its inverse. This coordinate equivalence has been discussed in past works \cite{SinclairNLD, ButcherDoubleTrans, Burnett2018_J2}. As a result of the linear equivalence, the relative motion problem can be explored from the linearized Cartesian perspective, if desired, then a linearized transformation to spherical coordinates followed by a nonlinear transformation back to Cartesian coordinates will reproduce the curvature correction offered by the spherical coordinate representation. However, the different modal representation in local spherical coordinates might offer benefits in some applications over the Cartesian representation. This is one of the topics explored in the numerical analysis in Section 3.4.

One interesting result from the preceding analysis is that the LTI form for spherical coordinates has a comprehensible physical interpretation with very simple dynamics. The LTI form for spherical coordinates is reproduced below, where $\bm{\chi}_{\bm{x}_{s}} = [P_{\bm{x}_{s}}(\theta)]^{-1}\bm{x}_{s}$ is the transformed state. 
\begin{equation}
\label{chiDynSph}
\bm{\chi}_{\bm{x}_{s}}' = [R_{\bm{x}_{s}}]\bm{\chi}_{\bm{x}_{s}}
\end{equation}
Examining the plant matrix, which is given by Eq. \eqref{RmatSph}, a simple interpretation of the dynamics in the spherical LTI coordinates is possible, because the three nonzero columns of $[R_{\bm{x}_{s}}]$ are linearly dependent. Factoring out $\alpha = 2R_{21}a/\gamma$, the following common column vector is defined:
\begin{equation}
\label{Rf}
\bm{R}_{f} = \alpha\begin{pmatrix} AC \\ \frac{C(B+1)^{2}}{\gamma a} \\ 0 \\ B \\ -\frac{2A(B+1)}{\gamma a} \\ 0 \end{pmatrix}
\end{equation}
The common column vector is related to all nonzero columns of $[R_{\bm{x}_{s}}]$ as below:
\begin{equation}
\label{RfToCol}
\bm{R}_{1} = \frac{B+2}{C}\bm{R}_{f}, \ \bm{R}_{4} = A\bm{R}_{f}, \ \bm{R}_{5} = \gamma a\bm{R}_{f}
\end{equation}
The relative motion state is resolved in the spherical LTI coordinates as the 6 component state vector $\bm{\chi}_{\bm{x}_{s}}$. The coordinates $\chi_{3}$ and $\chi_{6}$ are stationary -- see the zero 3rd and 6th rows of the plant matrix in Eq. \eqref{RmatSph}. Defining $\bm{\rho} = (\chi_{1}, \ \chi_{4}, \ \chi_{5})^{\top}$, the nonzero natural dynamics of the LTI coordinates are given below:
\begin{equation}
\label{sphLTI1}
\bm{\rho}' = \alpha\left(\bm{\rho}\cdot\bm{n}\right)\bm{\zeta}
\end{equation}
\begin{equation}
\label{sphLTI2}
\chi_{2}' = \alpha\frac{C(B+1)^{2}}{\gamma A}\left(\bm{\rho}\cdot\bm{n}\right)
\end{equation}
\begin{equation}
\label{sphLTI_n}
\bm{n} = \begin{pmatrix} \frac{B+2}{C} \\ A \\ \gamma a\end{pmatrix}
\end{equation}
\begin{equation}
\label{sphLTI_xi}
\bm{\zeta} = \begin{pmatrix} AC \\ B \\ -\frac{2A(B+1)}{\gamma a} \end{pmatrix}
\end{equation}
where $\bm{\zeta}\cdot\bm{n} = 0$ and $\frac{d}{d\theta}\left(\bm{\rho}\cdot\bm{n}\right) = 0$. From Eqs. \eqref{sphLTI1} and \eqref{sphLTI2}, it is clear that the dynamics of $\chi_{2}$ are influenced by $\bm{\rho}$, but the coordinate $\chi_{2}$ does not influence $\bm{\rho}$. The 3D space in which $\bm{\rho}$ is embedded is by far the most important space for the transformed relative motion problem. When $\bm{\rho}\cdot\bm{n} = 0$, the dynamics are stationary. This is an equation of a plane passing through $\bm{\rho} = \bm{0}$. This plane is called the stationary plane. The vector field of the dynamics of $\bm{\rho}$ is parallel to the stationary plane, pointing along $\bm{\zeta}$ above the plane and along $-\bm{\zeta}$ below it. The magnitude of the vector field at any point is proportional to the distance off the plane, $\bm{\rho}\cdot\bm{n}$.

The coordinates $\chi_{3}$ and $\chi_{6}$ are related to the out-of-plane motion, and they are decoupled from the in-plane motion and stationary. Any periodic in-plane motion of interest can be parameterized by a unique choice of four constant state values $\chi_{1}, \ \chi_{2}, \ \chi_{4}, \ \chi_{5}$, where $\bm{\rho} = (\chi_{1}, \ \chi_{4}, \ \chi_{5})^{\top}$ is constrained to the stationary plane to prevent movement of the $\chi_{2}$ coordinate. The dynamics of $\bm{\rho}$ and $\chi_{2}$ are easily described and visualized, as discussed above. By the LF transformation to the spherical LTI coordinates, the dimensionality of the Keplerian satellite relative motion problem is reduced to 3 active coordinates with very simple dynamics. The other coordinates are a steered coordinate $\chi_{2}$ and the stationary out-of-plane coordinates. 

Given the simplicity of the dynamics in the spherical LTI coordinates, a natural question is to explore control design in the context of this formulation. 
For example, low-thrust strategies for relative orbit reconfiguration will naturally maintain close proximity to the stationary plane, to minimize the degree to which the natural dynamics need to be countered by control. By contrast, impulsive maneuver-based strategies will push $\bm{\rho}$ further from the stationary plane and will make greater use of the natural dynamics to achieve control objectives. Examining existing control strategies from the perspective of this special coordinate representation might provide new insights.

\subsection{Numerical Simulations for the Keplerian Problem}
In this section, the developments in this paper are tested on unperturbed Earth orbits with $a = 26600 \ \text{km}$, $\Omega = 0^{\circ}$, $i = 63.4^{\circ}$, and $\omega = 270^{\circ}$, and the eccentricity is varied. In the case that $e \sim 0.74$, the resulting orbit is of the same type as the Molniya orbits used by the Soviets. 

\subsubsection{Modal Decomposition in Local Cartesian Coordinates}
\begin{figure}[h!]
\centering
\includegraphics[]{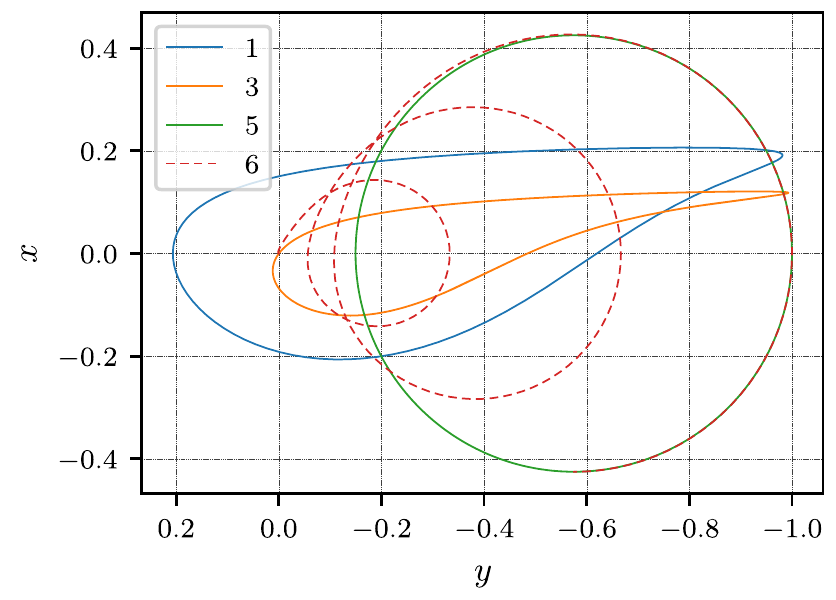}
\caption{In-plane Normalized Relative Motion Modes, $e = 0.74$, $f_{0} = 90^{\circ}$}
\label{fig:modeIntro2D}
\end{figure}

\begin{figure}[h!]
\centering
\includegraphics[]{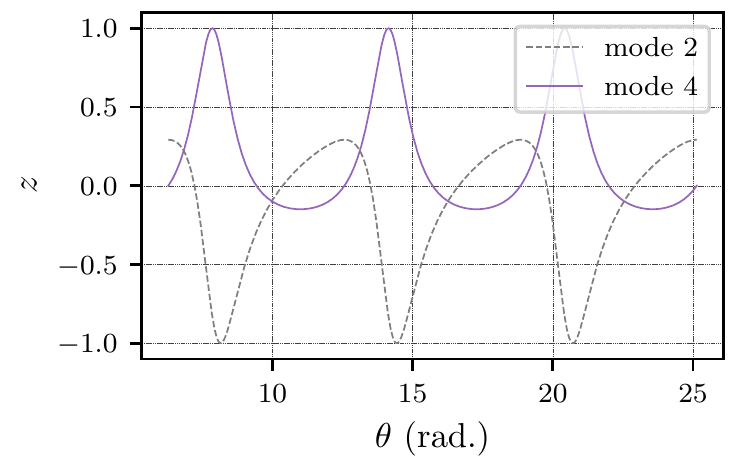}
\caption{Out-of-plane Normalized Relative Motion Modes, $e = 0.74$, $f_{0} = 90^{\circ}$}
\label{fig:modeIntroZ}
\end{figure}

The modal decomposition concept discussed in this paper enables any close-proximity relative motion to be expressed as the unique weighted combination of 6 or fewer modal motions. To introduce this concept, the 6 relative motion modes are computed for the Molniya orbit with $e = 0.74$. There are four in-plane modes (modes 1, 3, 5, 6), and two purely out-of-plane modes (2 and 4). These are normalized and plotted for three chief orbits in Figures \ref{fig:modeIntro2D} and  \ref{fig:modeIntroZ}, where the mode numbers correspond with the numbering of the constants in Eq. \eqref{c123}. 

\begin{figure}[h!]
\centering
\includegraphics[]{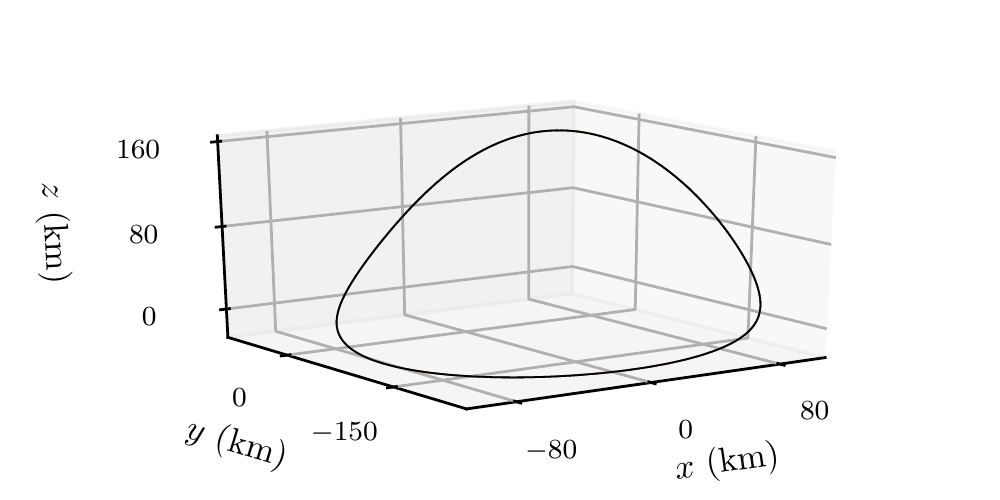}
\caption{Example Relative Orbit, $e = 0.74$, $f_{0} = 90^{\circ}$, $\delta e = 0.002$, $\delta i = 0.2^{\circ}$, $\delta f_{0} = 0^{\circ}$}
\label{fig:RO_molniya_ex1}
\end{figure}

The modes are numerically validated by comparing to propagation of their initial conditions with the Tschauner-Hempel equations and ensuring a match \cite{tschauner1}. All modes are periodic except the drift mode, mode 6. This mode is a composition of the motion of mode 5, but grows and drifts over time in the along-track direction. The 5th mode is an offset circle that is similar to a combination of fundamental solutions to the Tschauner-Hempel equations discussed in Reference \citenum{SinclairGeometricTH}. 

\begin{figure*}[h!]
\centering
\includegraphics[]{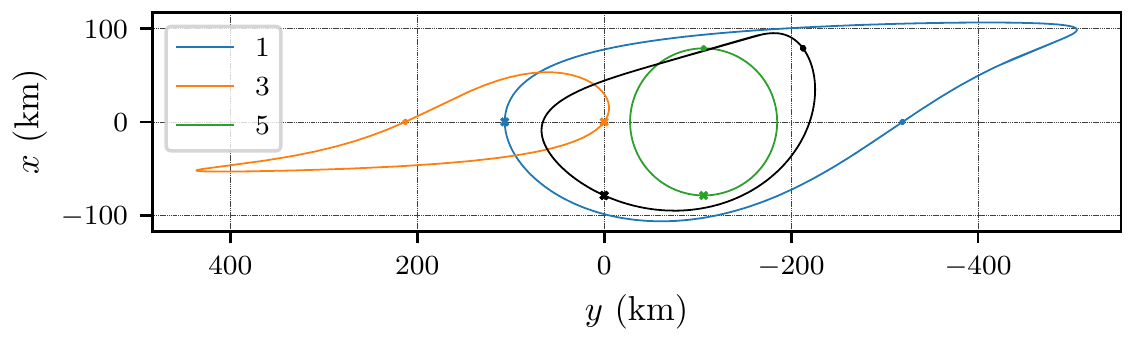}
\caption{Example Modal Decomposition (Planar), $e = 0.74$, $f_{0} = 90^{\circ}$, $\delta e = 0.002$, $\delta i = 0.2^{\circ}$}
\label{fig:RO_ex1_modal2D}
\end{figure*}

To demonstrate how these relative motion modes combine to construct any close-proximity relative motion, consider an example of bounded relative motion with $\delta a = 0$, $\delta e = 0.002$, and $\delta i = 0.2^{\circ}$. The resulting relative motion is depicted in Figure \ref{fig:RO_molniya_ex1}. This is distorted from the traditional 2:1 relative orbit ellipse of the Clohessy-Wiltshire solution, due to the very high chief eccentricity. Because the motion starts out in the $x$-$y$ plane, the out-of-plane motion is constructed entirely of mode 4. The in-plane motion is composed of three of the in-plane modes. Figure \ref{fig:RO_ex1_modal2D} shows the three modes that combine to construct the planar component of the relative motion, which is given in black. The modes are scaled such that their linear combination produces the relative orbit, and this result is numerically confirmed. Note the absence of any contribution of the drift mode, as expected. The initial position is marked with an \texttt{x}, and the point after a true anomaly change of $\Delta f = \pi$ is marked with a filled circle. Using these points, it is possible for the reader to graphically verify that the sum of the individual modes reproduces the indicated motion. Note from Figure \ref{fig:RO_ex1_modal2D} that $c_{3} < 0$, because mode 3 is flipped in comparison to its normalized form in Figure \ref{fig:modeIntro2D}.


\begin{figure*}[h!]
\centering
\includegraphics[]{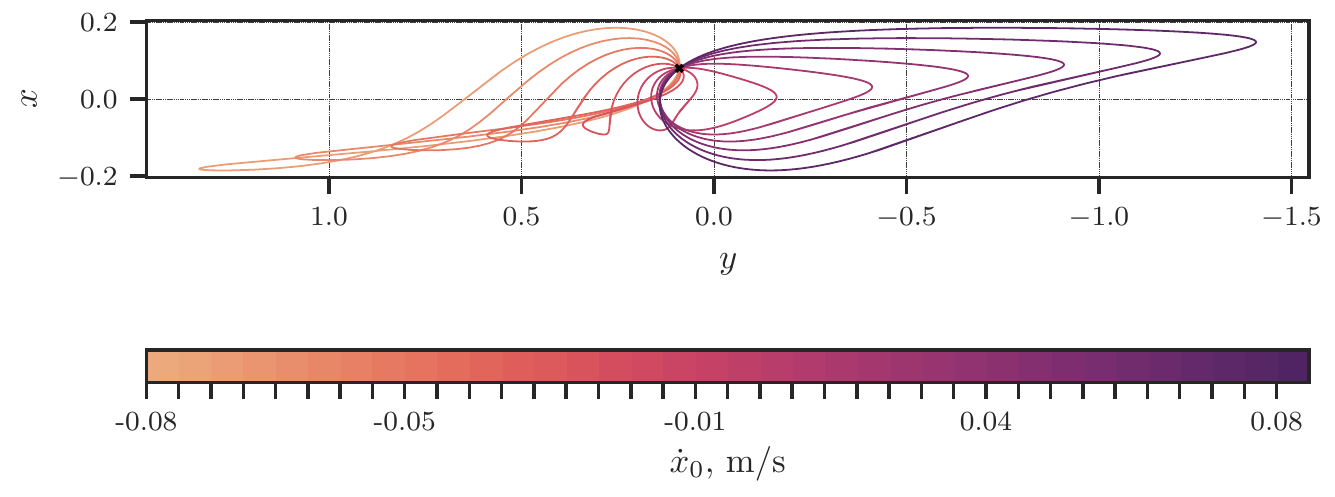}
\caption{Bounded Planar Motion with $(x_{0}, \ y_{0})$ specified, $e = 0.74$, $f_{0} = 90^{\circ}$}
\label{fig:RO_ex1_specvar1}
\end{figure*}

The analysis in this work enables analytic modal decomposition using Eqs. \eqref{LF_map1}, \eqref{VforRmatCart}, and \eqref{c123}. Furthermore, only Eq. \eqref{c123} needs to be re-evaluated for each possible relative motion case -- the eigenvectors of the LTI plant matrix and the periodic transformation only change with the chief orbit. In addition, Eq. \eqref{c123} is simple enough to facilitate some interesting analysis that leverages the computational efficiency of this formulation. For example, recall that if the degree of drift is specified via a fixed value of $c_{6}$, a select initial in-plane component of the position $(x_{0}, \ y_{0})$ forms a point of intersection of all possible in-plane relative motions in a one-parameter variation, based on the value of $\dot{x}_{0}$. To demonstrate this for the Molniya orbit, consider an initial planar relative position of $(x_{0}, \ y_{0}) = (0.08, 0.09)$ km, and the drift constant is set to $c_{6} = 0$ to explore only bounded relative motion solutions. The variation of $\dot{x}_{0}$ yields the family of possible planar motions originating at the specified point $(x_{0}, y_{0})$. The modes only need to be computed once using Eqs. \eqref{LF_map1} and \eqref{VforRmatCart}, while repeated evaluation of expressions derived from Eq. \eqref{c123} facilitates the computation of the families of relative orbits intersecting the point of interest. A subset of the possible relative orbits is computed and given in Figure \ref{fig:RO_ex1_specvar1}, with the initial point indicated by an \texttt{x}.

\begin{figure*}[ht!]%
    \centering
    \subfloat[$e = 0.01$]{{\includegraphics[]{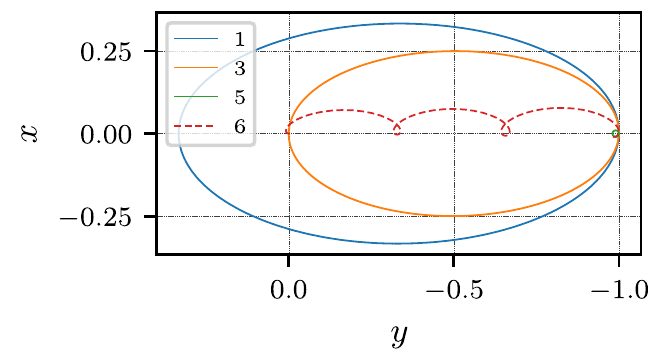} }}%
    \subfloat[$e = 0.1$]{{\includegraphics[]{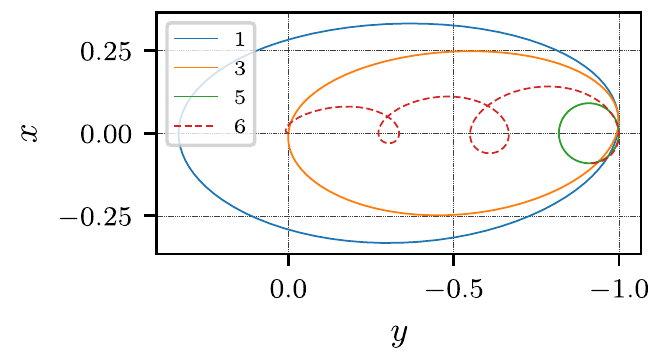} }}%
    \qquad
    \subfloat[$e = 0.5$]{{\includegraphics[]{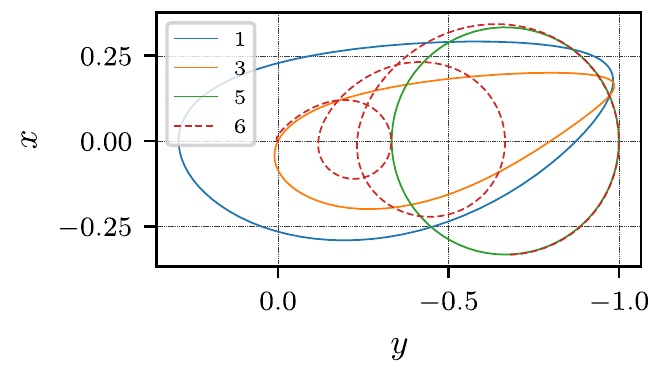} }}
    \subfloat[$e = 0.74$]{{\includegraphics[]{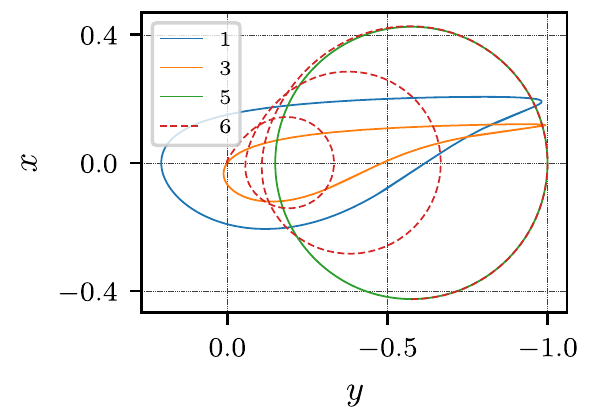} }}
    \caption{Normalized Relative Motion Modes vs. Eccentricity, Cartesian Coordinates}%
    \label{fig:modalCartEcc}%
\end{figure*}

In Figure \ref{fig:modalCartEcc}, the eccentricity is varied to show the evolution of the relative motion modes. The drift modes are plotted for three chief orbit periods. Because the model is linear, the scale of the modes is unimportant. All modes have been normalized in the figures so the maximum relative distance is unit magnitude. Starting with $e = 0.01$, the four planar modes are two relative motion ellipses, a drift mode, and a small circular motion in the along-track direction. While the formulation explored in this paper becomes singular for $e = 0$, it is still well-defined for small but nonzero values of eccentricity. The 2:1 centered relative motion ellipse from the classical relative motion problem with near-circular orbits would be constructed from modes 1, 3, and 5. Increasing the eccentricity to $0.1$, the circular mode becomes larger, and as a result, the loops in the drift mode also grow more noticeably over time. Increasing the eccentricity to $e = 0.5$, the first and third modes have become distorted. The nature of this distortion varies with the choice of epoch true anomaly $f_{0}$. For $e = 0.5$, mode 5 falls in the range $1/3 \leq x \leq 1$, with the center at $x = 2/3$. It is determined that mode 5 is a circle centered in the along-track direction, for which the eccentricity determines the ratio of the circle radius to the distance of its center from the origin. When the eccentricity is increased to that of the Molniya orbits, modes 1 and 3 become distorted significantly at their greatest along-track extent.

\subsubsection{Analysis using Local Spherical Coordinates}
\begin{figure*}[h!]%
    \centering
    \subfloat[Cartesian Coordinates]{{\includegraphics[]{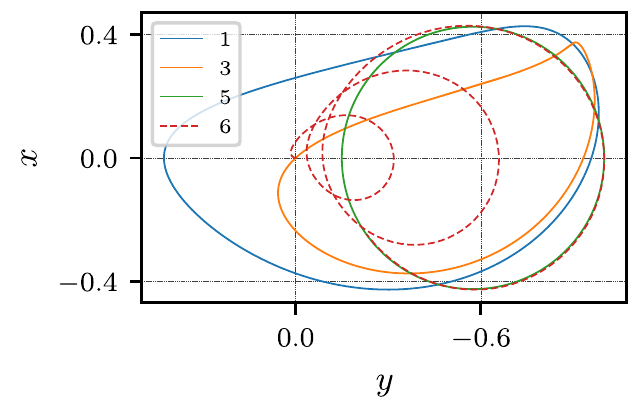} }}%
    \subfloat[Spherical Coordinates]{{\includegraphics[]{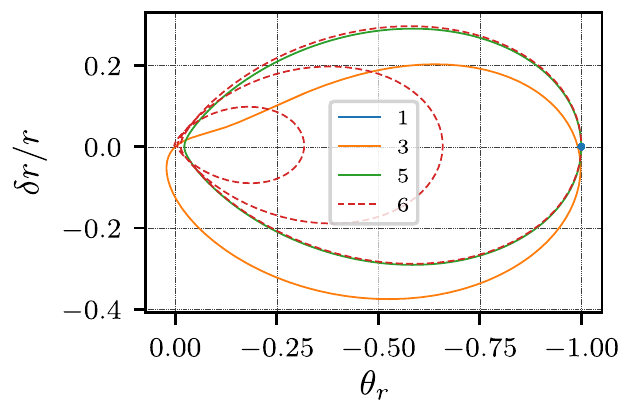} }}%
    \caption{Normalized Relative Motion Modes, $e = 0.74$, $f_{0} = 145^{\circ}$}%
    \label{fig:modalExTwo}%
\end{figure*}

Using the same Molniya orbit from the previous example, the epoch true anomaly is shifted to $f_{0} = 145^{\circ}$. The normalized planar modes from decompositions in Cartesian and spherical coordinates are provided in Figure \ref{fig:modalExTwo}. The Cartesian modes 1 and 3 are distorted differently from the $f_{0} = 90^{\circ}$ case, but modes 5 and 6 are still similar to before. In spherical coordinates, plotting $\delta r/r$ enables a visualization of the relative motion with the non-dimensional $\theta_{r}$ coordinate. Because $r$ varies greatly over time for sufficiently eccentric orbits, the motion plotted in the figure is not representative of the modal motion in Cartesian coordinates. For this reason, linearly mapping the spherical coordinate modes to Cartesian coordinates may be preferable for visualization. However, when plotted in these normalized spherical coordinates, the first mode becomes just a single point, which could be a useful simplification.

\begin{figure*}[h!]
\centering
\includegraphics[]{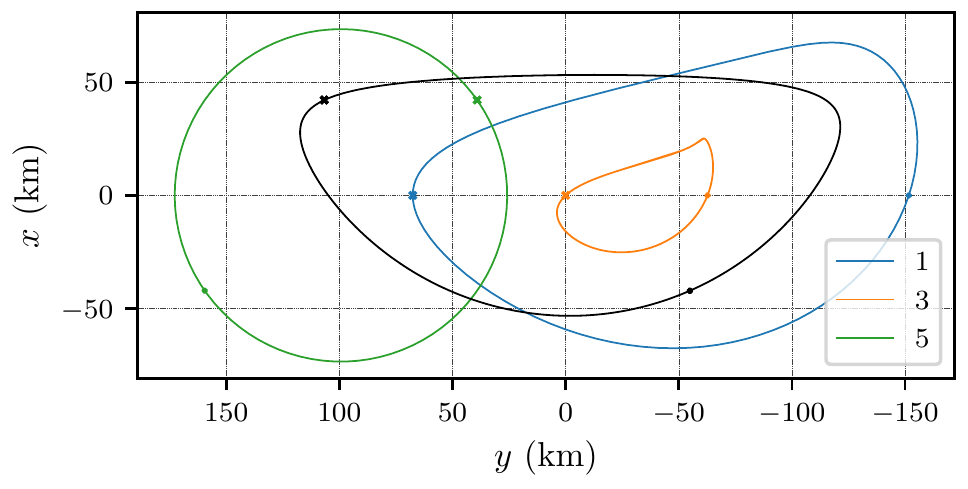}
\caption{Example Modal Decomposition, Cartesian, $e = 0.74$, $f_{0} = 145^{\circ}$, $\delta e = 0.002$, $\delta i = \delta f_{0} = 0.2^{\circ}$}
\label{fig:caseExTwoCart}
\end{figure*}

\begin{figure*}[h!]
\centering
\includegraphics[]{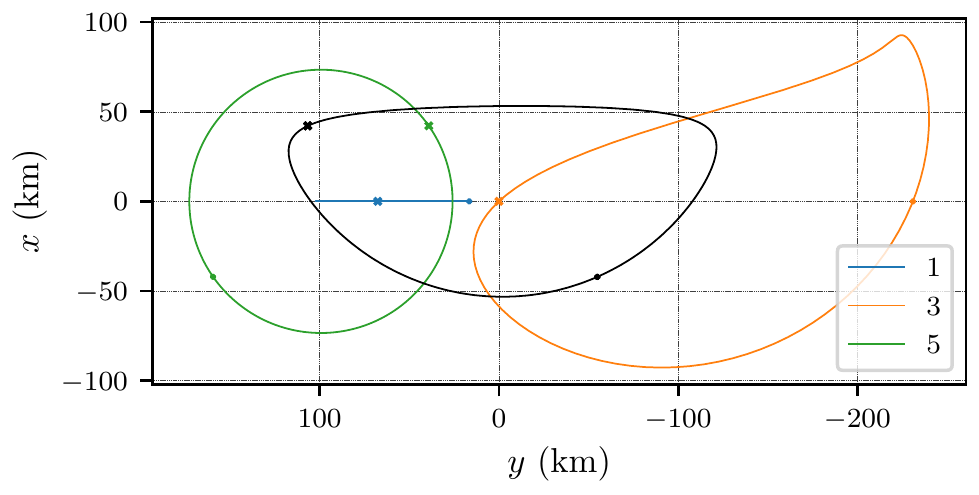}
\caption{Example Modal Decomposition, Spherical, $e = 0.74$, $f_{0} = 145^{\circ}$, $\delta e = 0.002$, $\delta i = \delta f_{0} = 0.2^{\circ}$}
\label{fig:caseExTwoSph}
\end{figure*}

A new relative motion example is parameterized by the initial orbit element differences $\delta e = 0.002$, $\delta i = \delta f_{0} = 0.2^{\circ}$. The planar component of the resulting relative orbit is expressed in terms of the Cartesian relative motion modes in Figure \ref{fig:caseExTwoCart}. The out-of-plane components of the motion are omitted from this analysis because they are comparatively uninteresting. In the same manner as the previous example, the initial point is marked with a \texttt{x} and the point after half an orbit at $\Delta f_{0} = 180^{\circ}$ is marked with a solid circle. In Figure \ref{fig:caseExTwoSph}, the same motion is plotted as a sum of the spherical coordinate modes expressed in Cartesian coordinates. As a result of this mapping, the stationary mode 1 in Figure \ref{fig:modalExTwo} becomes an oscillatory motion in the along-track direction. By inspection of the linear transformation in Eq. \eqref{CartToSphLin}, it is determined that the oscillatory along-track motion is due to rescaling by the chief orbit radius.

Comparing Figures \ref{fig:caseExTwoCart} and \ref{fig:caseExTwoSph}, the mapped spherical coordinate parameterization of the relative motion offers some simplifications over the Cartesian coordinate parameterization. While modes 3 and 5 are similar, mode 1 has been reduced to a simpler one-dimensional motion. With this parameterization, the manner in which modes 1, 3, and 5 linearly combine to produce the example relative motion is easier to visualize than using the Cartesian representation. There are only two 2D motions, and mode 1 shifts their sum in the along-track direction.

Through the application of the modal decomposition technique and the convenient mapping of the LF transform and LTI solutions across coordinates to the Keplerian relative motion problem, a few things have been demonstrated. First, it is shown that the choice of coordinates influences the geometric complexity of the modal solutions. From the technique of deriving new LF transformations from old, the exploration of relative motion modal decompositions derived from different coordinate representations is highly feasible. Additionally, simple conclusions have been obtained about the nature of relative motion in any bounded Keplerian orbit. Using the spherical coordinate modal decomposition, it is determined that any close-proximity natural relative motion can be expressed as the weighted sum of two purely out-of-plane modes, one in-plane drift mode, an offset circle mode, a 1D along-track oscillatory mode, and a ``teardrop" shaped mode. This simple basis of relative motion solutions facilitates straightforward design of close-proximity relative motion in Keplerian orbits of any eccentricity. The six modal constants parameterize all possible motions, similarly to how the six LROEs for the CW solutions \cite{Bennett2016c} can be used to explore all possible close-proximity relative motion for near-circular orbits.

\section{Application of the Theory to Perturbed Orbits}
In the case that the orbital dynamics are not Keplerian, the modal decomposition theory can still be applied if the chief orbits are sufficiently close to periodic. This section briefly discusses the procedure for analytic application, while also summarizing more advanced numerical application to a highly non-Keplerian environment in the vicinity of an asteroid. A complete derivation for a perturbed example is beyond the scope of this paper. 
\subsection{Analytic Application}
In the case of a sub-dominant non-Keplerian perturbation (such as $J_{2}$), the LTI matrix $[\Lambda]$, LF transformation $[P]$, and geometric transformation $[G]$ are all perturbed from their Keplerian forms $[\Lambda_{0}]$, $[P_{0}]$, and $[G_{0}]$:
\begin{subequations}
\label{PertMat_NS1}
\begin{align}
[\Lambda] = & \ [\Lambda_{0}] + [\delta\Lambda] \\
[P] = & \ [P_{0}] + [\delta P] \\
[G] = & \ [G_{0}] + [\delta G]
\end{align}
\end{subequations}
Solving for a first-order correction in the LF transformation $[\delta P]$ from the deviation in the plant matrix $[\delta A]$, the following differential equation for these quantities is obtained:
\begin{equation}
\label{PertMat_NS2}
[\delta\dot{P}] = -[\delta P][\Lambda_{0}] + [A_{0}][\delta P] - [P_{0}][\delta\Lambda] + [\delta A][P_{0}]
\end{equation}
Eq. \eqref{PertMat_NS2} is solved analytically using a specific $[\delta A]$ for the desired perturbative effects, analogously to how Eq.~\eqref{LFintro3} is solved. For this, it is easiest to solve for the LF transformation in the space of orbit element differences, because $[\delta A]$ will be much simpler in this space than it is in local coordinates. Additionally, as is done in Reference~\citenum{KoenigJ2}, it may be convenient to examine only the secular variations induced by the perturbations, ignoring short-period effects. This will significantly simplify $[\delta A]$. 
Lastly, the deviation $[\delta G]$ in the geometric mapping from orbit element differences to local coordinates must also be obtained. Reference~\citenum{gim1} discusses how this is obtained for the $J_{2}$ perturbation.

Once the above analysis has been performed, a modified modal decomposition will be obtained, with a modified vector of modal constants $\bm{c}$ that is still a function of relative state initial conditions. The same modal analysis as performed in this paper for the Keplerian problem can then be explored for the perturbed problem of interest. 
\subsection{Numerical Application}
Numerical application of the relative motion modal decomposition theory is easier to implement than analytic application. Explicit computation of the perturbed LF transformation and perturbed LTI plant matrix can be quite challenging, especially in the case of many perturbations. This sub-section summarizes a numerical application of the theory to examine motion in the vicinity of perturbed terminator orbits about the asteroid Ryugu.

Numerical application of the theory is as follows:
\begin{enumerate}
\item Numerically or symbolically linearize about the chief orbit in the desired coordinates to obtain $[A(t)]$.
\item If the chief orbit is periodic, skip to step 4. Otherwise, find the $T$ for which $[A(t_{0})] \approx [A(t_{0}+T)]$.
\item Compute the periodic part of $[A(t)]$, $[\overline{A}(t)]$, by Fourier series fit of the components of $[A(t)]$. The resulting $[\delta A(t)] = [A(t)] - [\overline{A}(t)]$ needs to be sub-dominant.
\item Numerically integrate an STM using $[\overline{A}]$,  obtain the monodromy matrix, then compute the resulting LTI matrix and LF transformation using Eqs.~\eqref{LFintro6} and \eqref{LFintro4}.
\item The fundamental modal solutions can be evaluated numerically using the procedure discussed in this paper.
\end{enumerate}
For non-periodic chief orbits, the accuracy of the modal solutions for describing all admissible close-proximity relative motion will depend on the relative scale of $[\delta A]$ and, relatedly, on the choice of $T$ \cite{Burnett2020_Ryugu}. 

The above numerical procedure has been applied to perturbed terminator orbits about the asteroid Ryugu. The results are summarized here.
\begin{figure}[h!]
\centering
\includegraphics[]{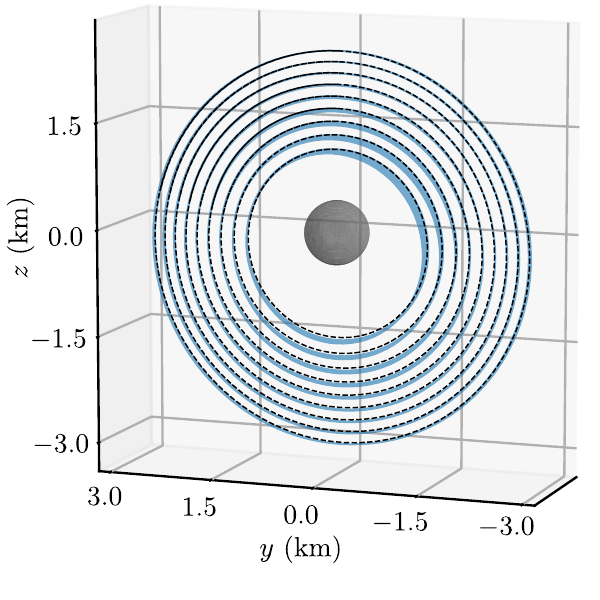}
\caption{Perturbed Terminator Orbits about Ryugu}
\label{fig:Terminator}
\end{figure}
These terminator orbits are found using the Augmented Normalized Hill Three-Body problem \cite{ScheeresMarzari2002} to compute the nominal initial conditions, and the resulting orbits are then perturbed from being exactly periodic by higher-order gravitational perturbations from the body, nonlinear solar gravity, and the effects of the asteroid's orbital eccentricity. However, the plant matrices $[A(t)]$ for the orbits are still almost periodic. To maximize periodicity of $[A(t)]$, a terminator orbit at an altitude yielding $\Gamma = T/T_{r} = 3$ is selected, for which the asteroid completes three rotations for every orbital revolution. 

\begin{figure}[h!]
\centering
\includegraphics[]{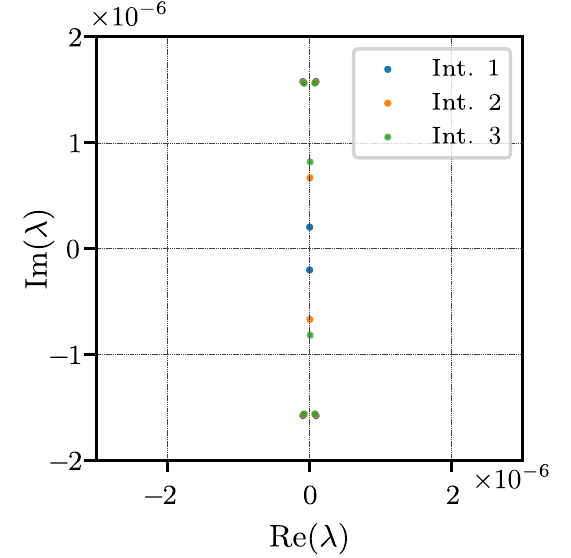}
\caption{Eigenvalues of the LTI Matrix -- Three Intervals, $\Gamma = 3$ Terminator orbit}
\label{fig:EVs_Term}
\end{figure}

\begin{figure}[h!]
\centering
\includegraphics[]{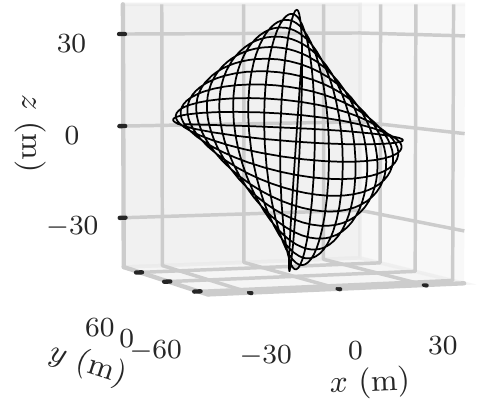}
\caption{Select Modal Motion (12 orbits)}
\label{fig:Modes_Ryugu1}
\end{figure}

The modal decomposition is computed, and the eigenvalues of the resulting LTI matrix are given in Figure \ref{fig:EVs_Term} for three subsequent orbits. Note that because $[A(t)]$ is not periodic, there is some drift over time in one of the computed modes -- indicated by the imaginary eigenvalue pair shifting over time. The other four eigenvalues are largely stationary, and those modes change only very slowly with each subsequent orbit. Figure \ref{fig:Modes_Ryugu1} gives the modal motion associated with the pair of eigenvalues with $\text{Re}(\lambda_{i,j}) > 0$. 
Simulations performed with and without the higher-order gravitational disturbances yield highly similar modes, indicating that those disturbances are sub-dominant for terminator orbits at this altitude, and don't greatly affect the characteristics of relative motion.
Additionally, simulations show that the motion in the vicinity of the terminator orbits is largely oscillatory, with the modes exhibiting complex shapes. The numerical application of the modal decomposition enables an approximate basis for admissible relative motions to be computed fairly efficiently for this highly complicated case.

As a test of the arguments in Section 4.1, the numerical modal decomposition procedure is applied to the $J_{2}$ problem in the space of QNS orbit element differences. In this case, numerical results show that the numerically computed LF transformation is a perturbed form of the Keplerian case (Eq.~\eqref{dtheta_sol2}), with smaller additional time-varying oscillations about zero in many components, represented by $[P_{\epsilon}(t)]$:
\begin{equation}
\label{NearIdentTrans1}
\begin{pmatrix} \delta a \\ \delta\theta \\ \delta i \\ \delta q_{1} \\ \delta q_{2} \\ \delta\Omega \end{pmatrix} \approx  \begin{bmatrix} 1 & 0 & 0 & 0 & 0 & 0 \\ 0 & P_{2,2}(t) & 0 &  P_{2,4}(t) &  P_{2,5}(t) & 0 \\ 0 & 0 & 1 & 0 & 0 & 0 \\ 0 & 0 & 0 & 1 & 0 & 0 \\ 0 & 0 & 0 & 0 & 1 & 0 \\ 0 & 0 & 0 & 0 & 0 & 1 \end{bmatrix}\begin{pmatrix} z_{1} \\ z_{2} \\ z_{3} \\ z_{4} \\ z_{5} \\ z_{6} \end{pmatrix} + [P_{\epsilon}(t)]\bm{z}
\end{equation}
The bottom row of $[P_{\epsilon}(t)]$ is determined to be zero, thus $\delta\Omega = z_{6}$, and furthermore the mode associated with the $z_{6}$ coordinate has a zero eigenvalue, recovering the invariance of relative motion in the $J_{2}$ problem with respect to $\delta\Omega$. This result highlights the possibility of future analytic work for computing the modal decompositions. 

Because the plant matrix is not exactly periodic for the asteroid and $J_{2}$ numerical examples, the computed LF transformation and the resulting modal decomposition will form an approximate basis of possible close-proximity relative motion, and not an exact one as in the Keplerian case. However, if the discarded component $[\delta A(t)]$ is sufficiently small, the errors will be quite small. In some cases, the plant matrix $[A(t)]$ will be quasi-periodic, and it could be worthwhile to explore computing the quasi-periodic Lyapunov-Perron (LP) transformation \cite{Jorba_reduceJDE} that reduces the linearized dynamics to an LTI form. There are practical challenges to reducing the linearized dynamic equations with quasi-periodic coefficients. The decomposition based on the LF transformation, however, is numerically straightforward, analytically promising, and provides insights into the types of relative motion for a large range of potential applications.

\section{Conclusion}
This paper explores the modal decomposition concept for efficient and convenient parameterization of the spacecraft relative motion problem in the vicinity of any closed orbit. This approach is facilitated by Lyapunov-Floquet theory, enabling the LTV dynamics to be transformed into an LTI system. This paper introduces a means for computing the LF transformation for the decomposition in any set of coordinates using the LF transformation from another set of coordinates and the linearized mapping between coordinates. The procedure is applied to the Keplerian relative motion problem to obtain modal decompositions in local Cartesian and spherical coordinates using the LF transformation in orbit element differences. The resulting decompositions are analyzed for relative motion near a Molniya orbit. A numerical example with terminator orbits shows that the modal concept extends beyond the Keplerian case.

This work connects strongly to concepts from literature. First, it demonstrates the connection between the relative motion solution in orbit element differences explored by Reference \citenum{schaub} and the concept of the LF transformation applied to those coordinates. Additionally, it makes use of a similar geometric method concept to what was used to great effect in Reference \citenum{gim1}. Finally, some of the modal solutions obtained are equivalent to previously explored special combinations of the Tschauner-Hempel fundamental solutions \cite{SinclairGeometricTH}. New results include the spherical coordinate modal decomposition, the numerically efficient exploration of bounded relative motion using the modal solutions, and the extension of the theory for perturbed settings.

The benefits of the modal relative motion perspective discussed by this paper are numerous. First, the modal solution constants $\bm{c}$ offer a simple state representation for relative motion that has clear geometric meaning through the associated modes, and allows for computationally efficient exploration of possible relative motion types. The dynamics of the modal solution constants are functions only of control and perturbations. In the case of nominal dynamics without control, they are integral quantities, similar to the stationary ROE quantities explored elsewhere in literature. However, unlike traditional ROEs, modal constants can be computed beyond the Keplerian problem. For example, in periodic orbits in three-body environments, it would still be possible to compute a modal decomposition with associated modal constants that are stationary in the absence of additional perturbations or control. Thus, the modal decomposition perspective is a unified view that extends from the simple Clohessy-Wiltshire case to periodic orbits in exotic environments, with practical application extending even to the case of almost-periodic orbits encountered in real-world scenarios.

\subsection*{Acknowledgements}
This work was supported by the U.S. Department of Defense through the NDSEG Fellowship Program.

\bibliographystyle{cas-model2-names}

\bibliography{references.bib}

\begin{thebibliography}{28}
\expandafter\ifx\csname natexlab\endcsname\relax\def\natexlab#1{#1}\fi
\providecommand{\url}[1]{\texttt{#1}}
\providecommand{\href}[2]{#2}
\providecommand{\path}[1]{#1}
\providecommand{\DOIprefix}{doi:}
\providecommand{\ArXivprefix}{arXiv:}
\providecommand{\URLprefix}{URL: }
\providecommand{\Pubmedprefix}{pmid:}
\providecommand{\doi}[1]{\href{http://dx.doi.org/#1}{\path{#1}}}
\providecommand{\Pubmed}[1]{\href{pmid:#1}{\path{#1}}}
\providecommand{\bibinfo}[2]{#2}
\ifx\xfnm\relax \def\xfnm[#1]{\unskip,\space#1}\fi
\bibitem[{Bennett and Schaub(2016)}]{Bennett2016c}
\bibinfo{author}{Bennett, T.}, \bibinfo{author}{Schaub, H.},
  \bibinfo{year}{2016}.
\newblock \bibinfo{title}{Continuous-time modeling and control using
  nonsingular linearized relative-orbit elements}.
\newblock \bibinfo{journal}{Journal of Guidance, Control, and Dynamics}
  \bibinfo{volume}{39}, \bibinfo{pages}{2605--2614}.
\newblock \URLprefix \url{https://doi.org/10.2514/1.G000366}.
\bibitem[{Burnett and Schaub(2020)}]{Burnett2020_Ryugu}
\bibinfo{author}{Burnett, E.}, \bibinfo{author}{Schaub, H.},
  \bibinfo{year}{2020}.
\newblock \bibinfo{title}{{Modal Decomposition of Spacecraft Relative Motion in
  Quasi-Periodic Orbits}}, in: \bibinfo{booktitle}{AAS/AIAA Astrodynamics
  Specialist Conference}, \bibinfo{organization}{American Astronautical
  Society}.
\bibitem[{Burnett et~al.(2018)Burnett, Butcher, Sinclair and
  Lovell}]{Burnett2018_J2}
\bibinfo{author}{Burnett, E.R.}, \bibinfo{author}{Butcher, E.A.},
  \bibinfo{author}{Sinclair, A.J.}, \bibinfo{author}{Lovell, T.A.},
  \bibinfo{year}{2018}.
\newblock \bibinfo{title}{{Linearized Relative Orbital Motion Model About an
  Oblate Body Without Averaging}, {AAS} 18-218}.
\newblock \bibinfo{journal}{Advances in the Astronautical Sciences}
  \bibinfo{volume}{167}, \bibinfo{pages}{691--710}.
\bibitem[{Butcher and Lovell(2016)}]{ButcherDoubleTrans}
\bibinfo{author}{Butcher, E.A.}, \bibinfo{author}{Lovell, T.A.},
  \bibinfo{year}{2016}.
\newblock \bibinfo{title}{{Spherical Coordinate Perturbation Solutions to
  Relative Motion Equations: Application to Double Transformation Spherical
  Solution}}, in: \bibinfo{booktitle}{AIAA/AAS Spaceflight Mechanics Meeting},
  \bibinfo{organization}{American Astronautical Society}.
  \bibinfo{publisher}{Univelt}, \bibinfo{address}{Napa, CA}. pp.
  \bibinfo{pages}{3455--3475}.
\bibitem[{Casotto(2016)}]{Casotto}
\bibinfo{author}{Casotto, S.}, \bibinfo{year}{2016}.
\newblock \bibinfo{title}{{The Equations of Relative Motion in the Orbital
  Reference Frame}}.
\newblock \bibinfo{journal}{Celestial Mechanics and Dynamical Astronomy}
  \bibinfo{volume}{124}, \bibinfo{pages}{215--234}.
\newblock \URLprefix \url{https://doi.org/10.1007/s10569-015-9660-1}.
\bibitem[{Clohessy and Wiltshire(1960)}]{clohessy1}
\bibinfo{author}{Clohessy, W.H.}, \bibinfo{author}{Wiltshire, R.S.},
  \bibinfo{year}{1960}.
\newblock \bibinfo{title}{{Terminal Guidance System for Satellite Rendezvous}}.
\newblock \bibinfo{journal}{Journal of the Aerospace Sciences}
  \bibinfo{volume}{27}, \bibinfo{pages}{653--658}.
\bibitem[{D'Amico and Montenbruck(2006)}]{DAmico_EccInc}
\bibinfo{author}{D'Amico, S.}, \bibinfo{author}{Montenbruck, O.},
  \bibinfo{year}{2006}.
\newblock \bibinfo{title}{Proximity operations of formation-flying spacecraft
  using an eccentricity/inclination vector separation}.
\newblock \bibinfo{journal}{Journal of Guidance, Control, and Dynamics}
  \bibinfo{volume}{29}, \bibinfo{pages}{554--563}.
\newblock \URLprefix \url{https://doi.org/10.2514/1.15114}.
\bibitem[{Deshmukh and Sinha(2004)}]{Dshmukh_Sinha_LFControl}
\bibinfo{author}{Deshmukh, V.S.}, \bibinfo{author}{Sinha, S.C.},
  \bibinfo{year}{2004}.
\newblock \bibinfo{title}{{Control of Dynamic Systems with Time-Periodic
  Coefficients via the Lyapunov-Floquet Transformation and Backstepping
  Technique}}.
\newblock \bibinfo{journal}{Journal of Vibration and Control}
  \bibinfo{volume}{10}, \bibinfo{pages}{1517--1533}.
\newblock \URLprefix \url{https://doi.org/10.1177/1077546304042064}.
\bibitem[{Gim and Alfriend(2003)}]{gim1}
\bibinfo{author}{Gim, D.W.}, \bibinfo{author}{Alfriend, K.T.},
  \bibinfo{year}{2003}.
\newblock \bibinfo{title}{The state transition matrix of relative motion for
  the perturbed non-circular reference orbit}.
\newblock \bibinfo{journal}{{AIAA} Journal of Guidance, Control, and Dynamics}
  \bibinfo{volume}{26}, \bibinfo{pages}{956--971}.
\newblock \URLprefix \url{https://doi.org/10.2514/2.6924}.
\bibitem[{Han et~al.(2019)Han, Chen, Alonso, Rao, Cubas, Yin and
  Wang}]{Han2019OEstoSph}
\bibinfo{author}{Han, C.}, \bibinfo{author}{Chen, H.}, \bibinfo{author}{Alonso,
  G.}, \bibinfo{author}{Rao, Y.}, \bibinfo{author}{Cubas, J.},
  \bibinfo{author}{Yin, J.}, \bibinfo{author}{Wang, X.}, \bibinfo{year}{2019}.
\newblock \bibinfo{title}{A linear model for relative motion in an elliptical
  orbit based on a spherical coordinate system}.
\newblock \bibinfo{journal}{Acta Astronautica} \bibinfo{volume}{157},
  \bibinfo{pages}{465 -- 476}.
\newblock \URLprefix
  \url{http://www.sciencedirect.com/science/article/pii/S009457651830910X}.
\bibitem[{Jorba and Sim{\'o}(1992)}]{Jorba_reduceJDE}
\bibinfo{author}{Jorba, {\`A}.}, \bibinfo{author}{Sim{\'o}, C.},
  \bibinfo{year}{1992}.
\newblock \bibinfo{title}{On the reducibility of linear differential equations
  with quasiperiodic coefficients}.
\newblock \bibinfo{journal}{Journal of Differential Equations}
  \bibinfo{volume}{98}, \bibinfo{pages}{111--124}.
\newblock \URLprefix
  \url{https://www.sciencedirect.com/science/article/pii/002203969290107X}.
\bibitem[{Kechichian(1998)}]{KechAngVel1998}
\bibinfo{author}{Kechichian, J.}, \bibinfo{year}{1998}.
\newblock \bibinfo{title}{Motion in general elliptic orbit with respect to a
  dragging and precessing coordinate frame}.
\newblock \bibinfo{journal}{Journal of Astronautical Sciences}
  \bibinfo{volume}{46}, \bibinfo{pages}{25 -- 45}.
\newblock \URLprefix \url{https://doi.org/10.1007/BF03546191}.
\bibitem[{Koenig et~al.(2017)Koenig, Guffanti and D'Amico}]{KoenigJ2}
\bibinfo{author}{Koenig, A.W.}, \bibinfo{author}{Guffanti, T.},
  \bibinfo{author}{D'Amico, S.}, \bibinfo{year}{2017}.
\newblock \bibinfo{title}{{New State Transition Matrices for Spacecraft
  Relative Motion in Perturbed Orbits}}.
\newblock \bibinfo{journal}{Journal of Guidance, Control, and Dynamics}
  \bibinfo{volume}{40}, \bibinfo{pages}{1749--1768}.
\newblock \URLprefix \url{https://doi.org/10.2514/1.G002409}.
\bibitem[{Lanczos(1986)}]{lanczos1986}
\bibinfo{author}{Lanczos, C.}, \bibinfo{year}{1986}.
\newblock \bibinfo{title}{The Variational Principles of Mechanics}.
\newblock Dover Books On Physics, \bibinfo{publisher}{Dover Publications}.
\newblock \URLprefix \url{https://books.google.com/books?id=ZWoYYr8wk2IC}.
\bibitem[{Meirovitch(2001)}]{meirovitch2001fundamentals}
\bibinfo{author}{Meirovitch, L.}, \bibinfo{year}{2001}.
\newblock \bibinfo{title}{Fundamentals of Vibrations}.
\newblock McGraw-Hill higher education, \bibinfo{publisher}{McGraw-Hill}.
\newblock \URLprefix \url{https://books.google.com/books?id=u358QgAACAAJ}.
\bibitem[{Montagnier et~al.(2004)Montagnier, Spiteri and
  Angeles}]{Montagnier_LFControl}
\bibinfo{author}{Montagnier, P.}, \bibinfo{author}{Spiteri, R.J.},
  \bibinfo{author}{Angeles, J.}, \bibinfo{year}{2004}.
\newblock \bibinfo{title}{{The control of linear time-periodic systems using
  Floquet--Lyapunov theory}}.
\newblock \bibinfo{journal}{International Journal of Control}
  \bibinfo{volume}{77}, \bibinfo{pages}{472--490}.
\newblock \URLprefix \url{https://doi.org/10.1080/00207170410001667477}.
\bibitem[{Nayfeh and Mook(1979)}]{Nayfeh:1979uq}
\bibinfo{author}{Nayfeh, A.H.}, \bibinfo{author}{Mook, D.T.},
  \bibinfo{year}{1979}.
\newblock \bibinfo{title}{Nonlinear Oscillations}.
\newblock \bibinfo{publisher}{John Wiley {\&} Sons, Inc.},
  \bibinfo{address}{New York, USA}.
\bibitem[{Ogundele and Agboola(2020)}]{OgundeleLFConference}
\bibinfo{author}{Ogundele, A.D.}, \bibinfo{author}{Agboola, O.A.},
  \bibinfo{year}{2020}.
\newblock \bibinfo{title}{Development of approximate solution of lf
  transformation of spacecraft relative motion with periodic coefficients}, in:
  \bibinfo{booktitle}{AAS/AIAA Astrodynamics Specialist Conference},
  \bibinfo{organization}{American Astronautical Society}.
\bibitem[{Schaub and Alfriend(2001a)}]{MeanElemImpulsive}
\bibinfo{author}{Schaub, H.}, \bibinfo{author}{Alfriend, K.T.},
  \bibinfo{year}{2001}a.
\newblock \bibinfo{title}{Impulsive feedback control to establish specific mean
  orbit elements of spacecraft formations}.
\newblock \bibinfo{journal}{{AIAA} Journal of Guidance, Control, and Dynamics}
  \bibinfo{volume}{24}, \bibinfo{pages}{739--745}.
\newblock \URLprefix \url{https://doi.org/10.2514/2.4774}.
\bibitem[{Schaub and Alfriend(2001b)}]{MeanElem}
\bibinfo{author}{Schaub, H.}, \bibinfo{author}{Alfriend, K.T.},
  \bibinfo{year}{2001}b.
\newblock \bibinfo{title}{{$J_2$ Invariant Relative Orbits for Spacecraft
  Formations}}.
\newblock \bibinfo{journal}{Celestial Mechanics and Dynamical Astronomy}
  \bibinfo{volume}{79}, \bibinfo{pages}{77--95}.
\newblock \URLprefix \url{https://doi.org/10.1023/A:1011161811472}.
\bibitem[{Schaub and Alfriend(2002)}]{LocalCart}
\bibinfo{author}{Schaub, H.}, \bibinfo{author}{Alfriend, K.T.},
  \bibinfo{year}{2002}.
\newblock \bibinfo{title}{Hybrid cartesian and orbit element feedback law for
  formation flying spacecraft}.
\newblock \bibinfo{journal}{{AIAA} Journal of Guidance, Control, and Dynamics}
  \bibinfo{volume}{25}, \bibinfo{pages}{387--393}.
\newblock \URLprefix \url{https://doi.org/10.2514/2.4893}.
\bibitem[{Schaub and Junkins(2018)}]{schaub}
\bibinfo{author}{Schaub, H.}, \bibinfo{author}{Junkins, J.L.},
  \bibinfo{year}{2018}.
\newblock \bibinfo{title}{Analytical Mechanics of Space Systems}.
\newblock \bibinfo{edition}{4th} ed., \bibinfo{publisher}{{AIAA} Education
  Series}, \bibinfo{address}{Reston, VA}.
\bibitem[{Scheeres and Marzari(2002)}]{ScheeresMarzari2002}
\bibinfo{author}{Scheeres, D.J.}, \bibinfo{author}{Marzari, F.},
  \bibinfo{year}{2002}.
\newblock \bibinfo{title}{Spacecraft dynamics in the vicinity of a comet}.
\newblock \bibinfo{journal}{Journal of the Astronautical Sciences}
  \bibinfo{volume}{50}, \bibinfo{pages}{35--52}.
\newblock \URLprefix \url{https://doi.org/10.1007/BF03546329}.
\bibitem[{Sherrill et~al.(2015)Sherrill, Sinclair, Sinha and
  Lovell}]{Sherrill2015LyapunovFloquet}
\bibinfo{author}{Sherrill, R.E.}, \bibinfo{author}{Sinclair, A.J.},
  \bibinfo{author}{Sinha, S.C.}, \bibinfo{author}{Lovell, T.A.},
  \bibinfo{year}{2015}.
\newblock \bibinfo{title}{{Lyapunov-Floquet control of satellite relative
  motion in elliptic orbits}}.
\newblock \bibinfo{journal}{IEEE Transactions on Aerospace and Electronic
  Systems} \bibinfo{volume}{51}, \bibinfo{pages}{2800--2810}.
\newblock \URLprefix \url{doi.org/10.1109/TAES.2015.140281}.
\bibitem[{Sinclair(2017)}]{SinclairNLD}
\bibinfo{author}{Sinclair, A.J.}, \bibinfo{year}{2017}.
\newblock \bibinfo{title}{{Calibrated and Decalibrated Approximations of
  Nonlinear Dynamic Systems}}.
\newblock \bibinfo{journal}{Nonlinear Dynamics} \bibinfo{volume}{87},
  \bibinfo{pages}{281--290}.
\newblock \URLprefix \url{https://doi.org/10.1007/s11071-016-3042-4}.
\bibitem[{Sinclair et~al.(2015)Sinclair, Sherrill and
  Lovell}]{SinclairGeometricTH}
\bibinfo{author}{Sinclair, A.J.}, \bibinfo{author}{Sherrill, R.E.},
  \bibinfo{author}{Lovell, T.A.}, \bibinfo{year}{2015}.
\newblock \bibinfo{title}{Geometric interpretation of the tschauner--hempel
  solutions for satellite relative motion}.
\newblock \bibinfo{journal}{Advances in Space Research} \bibinfo{volume}{55},
  \bibinfo{pages}{2268 -- 2279}.
\newblock \URLprefix
  \url{http://www.sciencedirect.com/science/article/pii/S0273117715000848}.
\bibitem[{Tschauner and Hempel(1965)}]{tschauner1}
\bibinfo{author}{Tschauner, J.}, \bibinfo{author}{Hempel, P.},
  \bibinfo{year}{1965}.
\newblock \bibinfo{title}{Rendezvous zu einem in elliptischer bahn umlaufenden
  ziel}.
\newblock \bibinfo{journal}{Astronautica Acta} \bibinfo{volume}{11},
  \bibinfo{pages}{104--109}.
\bibitem[{Yamanaka and Ankersen(2002)}]{yamanaka1}
\bibinfo{author}{Yamanaka, K.}, \bibinfo{author}{Ankersen, F.},
  \bibinfo{year}{2002}.
\newblock \bibinfo{title}{{New State Transition Matrix for Relative Motion on
  an Arbitrary Elliptical Orbit}}.
\newblock \bibinfo{journal}{{AIAA} Journal of Guidance, Control, and Dynamics}
  \bibinfo{volume}{25}, \bibinfo{pages}{60 -- 66}.
\newblock \URLprefix \url{https://doi.org/10.2514/2.4875}.

\end{thebibliography}


%

\end{document}